%% file: main.tex
\let\mypdfximage\pdfximage
\def\pdfximage{\immediate\mypdfximage}

\documentclass[10pt,twocolumn,letterpaper]{article}

\usepackage{cvpr}

\makeatletter
\@namedef{ver@everyshi.sty}{}
\makeatother

\usepackage{graphicx}
\usepackage{amsmath}
\usepackage{amssymb}
\usepackage{booktabs}
\usepackage{subcaption}
\usepackage{microtype}
\usepackage[normalem]{ulem}
\usepackage{adjustbox}
\usepackage{multirow}
\usepackage{tabularx}
\usepackage{makecell}
\usepackage{etoc}
\usepackage[accsupp]{axessibility}
\usepackage{xspace}
\usepackage{diagbox}
\usepackage[dvipsnames, table]{xcolor}

\usepackage{pgfplots, pgfplotstable}

\usepackage{mdframed}

\usepackage[pagebackref,breaklinks,colorlinks,citecolor=ForestGreen]{hyperref}

\usepackage[inline, shortlabels]{enumitem}

\usepackage[capitalize]{cleveref}

\input{tex/_abbrev_and_macros}
\input{tex/_pgf_abbrev}

\etocdepthtag.toc{mtchapter}
\etocsettagdepth{mtappendix}{none}

\begin{document}

\title{Fake it till you make it: \\
Learning transferable representations from synthetic ImageNet clones}

\author{
    Mert Bulent Sariyildiz$^{1,2}$
    \and
    Karteek Alahari$^2$
    \and
    Diane Larlus$^1$
    \and
    Yannis Kalantidis$^1$ \and \\
    $^1$~{NAVER LABS Europe} \hspace{1cm} $^2$~{Univ.\ Grenoble Alpes, Inria, CNRS, Grenoble INP, LJK}
}

\maketitle

\input{tex/0_abstract}
\input{tex/1_introduction}
\input{tex/2_related}
\input{tex/3_preliminaries}
\input{tex/4_method}
\input{tex/5_experiments}
\input{tex/6_discussion}
\input{tex/7_conclusions}

\vspace{-8pt}
\paragraph{Acknowledgements.}
This work was supported in part by MIAI@Grenoble Alpes (ANR-19-P3IA-0003), and the ANR grant AVENUE (ANR-18-CE23-0011).

{\small
\bibliographystyle{ieee_fullname}
\bibliography{main}
}

\appendix

\etocdepthtag.toc{mtappendix}
\etocsettagdepth{mtappendix}{subsection}
\etocsettagdepth{mtchapter}{none}

{
  \hypersetup{linkcolor=black}
  \tableofcontents
}
\input{tex/99_supplementary}

\end{document}

%% file: tex/_abbrev_and_macros.tex
\newcommand{\imnethlong}{ImageNet-100\xspace}

\newcommand{\imnet}{IN1K\xspace}
\newcommand{\imnetlong}{ImageNet-1K\xspace}

\newcommand{\imnethsd}{IN100-SD\xspace}
\newcommand{\imnethsdlong}{ImageNet-100-SD\xspace}

\newcommand{\imnetsd}{IN1K-SD\xspace}
\newcommand{\imnetsdlong}{ImageNet-1K-SD\xspace}

\newcommand{\diffup}[1]{{\color{Green}{($\uparrow$ #1)}}}

\newcommand{\colorname}{Brick-colored}
\newcommand{\better}[1]{{\color{BrickRed}{#1}}}

\newcommand\myrowno[1]{%
    \parbox{0.45cm}{{\color{gray} \scriptsize \hfill #1}}~~%
}

\newcommand{\sectioncolor}{Violet}
\mdfdefinestyle{DatasheetFrame}{
    linecolor=\sectioncolor,
    outerlinewidth=2pt,
    innertopmargin=4pt,
    innerbottommargin=4pt,
    innerrightmargin=4pt,
    innerleftmargin=4pt,
        leftmargin = 2pt,
        rightmargin = 2pt
}

\newcommand{\classset}{\mathcal{C}}
\newcommand{\nclasses}{N}

\newcommand{\mypar}[1]{\vspace{0.00cm}\noindent {\bf #1.}}
\newcommand{\myparnodot}[1]{\vspace{0.00cm}\noindent {\bf #1}}

\newcommand{\vx}{\mathbf{x}}
\newcommand{\vy}{\mathbf{y}}
\newcommand{\vz}{\mathbf{z}}

\crefname{section}{Sec.}{Secs.}
\Crefname{section}{Sec.}{Secs.}
\Crefname{table}{Tab.}{Tabs.}
\crefname{table}{Tab.}{Tabs.}
\Crefname{figure}{Fig.}{Figs.}
\crefname{figure}{Fig.}{Figs.}

\newcommand\setrow[1]{\gdef\rowmac{#1}#1\ignorespaces}
\newcommand\clearrow{\global\let\rowmac\relax}

%% file: tex/_pgf_abbrev.tex
\usepackage{pgfplots, pgfplotstable}
\pgfplotsset{compat=newest}
\usepgfplotslibrary{fillbetween}

\newcommand{\realc}{MidnightBlue}
\newcommand{\sdc}{WildStrawberry}

\newcommand{\leg}[1]{\addlegendentry{#1}}

\tikzset{every mark/.append style={solid}}
\pgfplotsset{
	grid=both, width=\linewidth, try min ticks=5,
    legend cell align=left, 
    legend style={fill opacity=0.8},
	ylabel near ticks,
    xlabel near ticks,
    every tick label/.append style={font=\footnotesize},
}

\pgfplotsset{
    real1/.style={thick, color=\realc, mark=o, mark size=2pt},
    sd1/.style={thick, color=\sdc, mark=*, mark size=2pt},
    sd1t/.style={thick, dashed, color=\sdc, mark=triangle*, mark size=2pt},
    real5/.style={thick, dashed, color=\realc, mark=o, mark size=2pt},
    sd5/.style={thick, dashed, color=\sdc, mark=*, mark size=2pt},
}

%% file: tex/0_abstract.tex
\begin{abstract}
\looseness=-1
Recent image generation models such as Stable Diffusion have exhibited an impressive ability to generate fairly realistic images starting from a simple text prompt.
Could such models render real images obsolete for training image prediction models? In this paper, we answer part of this provocative question by investigating the need for real images when training models for ImageNet classification.
Provided only with the class names that have been used to build the dataset, we explore the ability of Stable Diffusion to generate synthetic clones of ImageNet and measure how useful these are for training classification models from scratch.
We show that with minimal and class-agnostic prompt engineering, ImageNet clones are able to close a large part of the gap between models produced by synthetic images and models trained with real images, for the several standard classification benchmarks that we consider in this study.
More importantly, we show that models trained on synthetic images exhibit strong generalization properties and perform on par with models trained on real data for transfer.

\noindent Project page: {\href{https://europe.naverlabs.com/imagenet-sd/}{https://europe.naverlabs.com/imagenet-sd/}}
\end{abstract}

%% file: tex/1_introduction.tex
\vspace{-5pt}

\section{Introduction}
\label{sec:introduction}

\input{fig/teaser.tex}

\looseness=-1

The rise of (shallow) machine learning~\cite{chen2001one,vedaldi2009multiple}
and later deep learning~\cite{krizhevsky2012alexnet,szegedy2015going,he2016resnet}
has entirely changed the landscape of computer vision research over the past few decades, shifting some of the focus from \textit{methods} to the \textit{training data} itself.
Datasets, initially of hundreds of images and dozens of classes~\cite{li2004caltech101,everingham2009pascal}, have grown in size and complexity,
and started becoming contributions in their own right.
They have been fueling the progress of computer vision
as much as, if not more than, the methods themselves.
ImageNet~\cite{deng2009imagenet}, and mainly its \imnetlong~\cite{russakovsky2015ilsvrc} subset of about 1 million annotated images, has impacted the field in an unprecedented way.
Yet, curating and annotating such a dataset comes at a very high money and labor cost.

\looseness=-1
The last couple of years have seen the rise of large and generic models, trained
on data which is less curated but orders of magnitude larger.
Those proved to be easily applicable, either directly, or combined with a tailored model, to a wide range of computer vision transfer tasks~\cite{radford2021clip,jia2021scaling,ilharco_gabriel_2021_5143773}.
They have also been used beyond prediction tasks, \eg, for text-conditioned image generation.
Models such as DALL-E~\cite{ramesh2021zero} or Stable Diffusion~\cite{rombach2022high} have demonstrated impressive image generation ability.
They produce fairly realistic synthetic images and exhibit
a high degree of compositionality.

Such generative models are trained on billion-scale datasets~\cite{schuhmann2022laion} composed of noisy image-text pairs scraped from the internet.
Although training such models is out of reach for most institutions, a few of them have been made available to the community.
Given the remarkable
ability of these generative models, it is only natural to ask provocative questions such as: \textit{Is there still a need for real images when training image prediction models?}

In this paper we explore this question through one of the most iconic computer vision datasets, ImageNet~\cite{deng2009imagenet}.
We study to which extent this dataset can be entirely replaced by synthetic images when learning deep models.
For this, we assume that we are provided with a set of classes,
and the Stable Diffusion \cite{rombach2022high} model
a generator that can produce realistic images from a textual prompt.

Our task is to learn an image classification model \textit{from scratch} using a dataset composed only of synthetic images.
We then evaluate the performance of this model on several datasets.
First and foremost, we measure how well
models and classifiers trained only on synthetic images
recognize
the training classes
in real images from the standard ImageNet validation set.
Then, we evaluate them on common datasets that test their resilience to domain shifts or adversarial examples, still for the ImageNet training classes.
Finally, we consider several transfer learning scenarios where we measure the generalization performance of our models to novel classes.
\Cref{fig:teaser} summarizes the main results by comparing models trained on two equally sized set of images from the same set of classes, one real and one synthetic, on a number of these tasks. The gap is surprisingly narrow, especially for some of these scenarios.

\looseness=-1
To summarize, our contributions are threefold.
First, we leverage Stable Diffusion~\cite{rombach2022high} and generate synthetic ImageNet clones, \ie, datasets with synthetic images for the ImageNet classes, using
class names as prompts.
We analyse the generated images, highlight important issues, and propose class-agnostic alterations to the basic prompt that reduce semantic issues and increase diversity.
Second, we train classification models using different ImageNet clones and
show that
they can achieve $91.7\%$ and $70.3\%$ top-5 accuracy on \imnethlong and \imnetlong respectively.
Finally, we evaluate the generalization capacity of our models.
We show that their performance gap with models trained on real images is reduced when testing for resilience to domain shifts or adversarial examples.
Moreover, we show that our models perform on par with models trained conventionally when testing on 15 transfer datasets.

%% file: fig/teaser.tex
\begin{figure}[t!]
    \centering
    \includegraphics[width=1\linewidth]{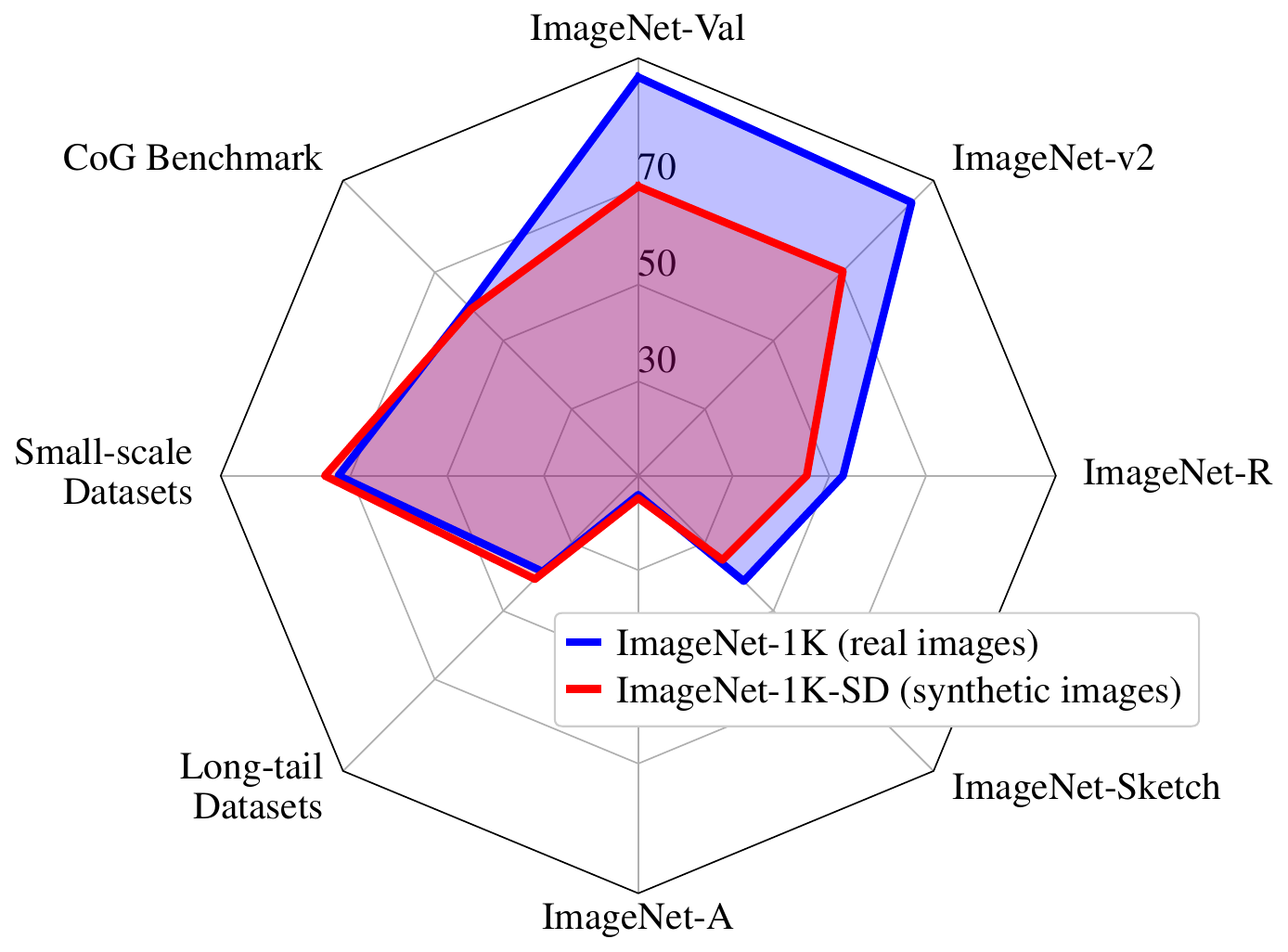}
    \caption{
        \textbf{\imnetlong vs ImageNet-1K-SD.}
        The {{\color{Blue}{blue polygon}}} shows the performance of a model trained on \imnetlong.
        The {{\color{Red}{red polygon}}} depicts the performance of one trained on ImageNet-1K-SD, \ie, {only on synthetic data} generated with Stable Diffusion~\cite{rombach2022high} using the class names of \imnetlong.
        We report top-5 accuracy for ImageNet test sets, and average top-1 for transfer tasks.
    }
    \label{fig:teaser}
\end{figure}

%% file: tex/2_related.tex
\section{Related work}
\label{sec:related}

\subsection{Learning with synthetic data}

\noindent
Learning with synthetic data has 
become a standard way to create large amounts of labeled data for annotation heavy tasks, 
such as human understanding \cite{varol2017learning,Pumarola_2019_ICCV},
semantic segmentation~\cite{sankaranarayanan2018learning,chen2019learning}, 
optical flow estimation~\cite{dosovitskiy2015flownet,kim2022how} or 
dense visual alignment~\cite{peebles2022gan}.
In most cases, this synthetic data requires access to 
3D models and renderers~\cite{mahmood2019amass}, or 
to a simulator~\cite{richter2016playing}
with a physically plausible engine.
Recent works propose pretraining on a database of synthetic fractal~\cite{KataokaIJCV2022} or sinusoidal wave~\cite{takashima2023visual} images before fine-tuning the model using real images on a downstream task.
In this study we use synthetic data to learn encoders and classifiers that can be used \textit{out-of-the-box}, without the need for a subsequent fine-tuning step.
Closest to our work, Kumar \etal~\cite{sreejith22evaluation} generate synthetic OCT images to train a glaucoma detection model to be applied to real images.
Here, we target synthetic clones of complex natural image datasets, \ie, \imnetlong~\cite{russakovsky2015ilsvrc}, and we use a {\em general-purpose} text-to-image generation model.

\mypar{Synthetic ImageNet clones}
Synthetic images for ImageNet classes have been used recently
in a number of related works~\cite{ravuri2019classification,besnier2020dataset,li2022bigdatasetgan} 
based on class conditional Generative Adversarial Networks (GANs), such as
BigGAN~\cite{brock2019large}.
Besnier \etal~\cite{besnier2020dataset} generate images for ten ImageNet classes and propose techniques to reduce the gap between models trained on generated images and real ones.
Li \etal~\cite{li2022bigdatasetgan} synthesize five images for
each ImageNet-1K class, together with their semantic segmentation annotations
to automatically generate pixel-level labels
at scale.
Our work focuses on image-level classification,
and uses a general-purpose text-conditioned generative model instead of ImageNet-1K class-conditioned GANs.
It further offers a larger scale study with promising results on the full ImageNet-1K benchmark
when training from 1.28 million synthetic images. 
Concurrent work~\cite{he2022synthetic} also synthesizes data for ImageNet-1K, but focuses on improvements on top of the CLIP~\cite{radford2021clip} model or after fine-tuning.

\mypar{Synthetic images as data++}
Data sampled from generative models~\cite{goodfellow2020generative,ramesh2021zero,ho2020denoising,rombach2022high} 
can be seen as data with added functionalities or ``data++''~\cite{isola2022when}.
Such data can be  manipulated, interpolated or composed
\cite{Jahanian2020on,chai2021ensembling,chai2021using,jahanian2022generative} with
dedicated operators in their latent
space, and further used for counterfactual reasoning~\cite{oktay2018counterfactual,liu2019generative,mao2021generative}.
In this paper, we
do not exploit these added functionalities. Our prompts consider a class at a time and do not leverage any interpolation nor the composition properties of synthetic data.
Instead, we chose our complete pipeline, including the set of data augmentations, to be identical to the one we use for real images, to allow for a fair comparison.

\looseness=-1
\mypar{Zero-shot learning and test-time view synthesis}
Generative models have been used to extend models to new classes, or to create novel views at test time.
Chai \etal~\cite{chai2021ensembling}
synthesize novel views for test-time ensembling by perturbing the latent code of a test image.
Aiming at zero-shot recognition~\cite{xian2018zero},
Elhoseiny \etal~\cite{elhoseiny2013write} synthesize a classifier for any novel class given its semantic description (\eg, textual or attribute-based),
whereas others synthesize 
images~\cite{dunlap2023using,gu2022can}, or image \textit{features}~\cite{sariyildiz2019gmn,lazarou2022tensor} using such descriptions.
Here we aim to learn encoders from scratch,
and do not rely on models previously trained on real data.

\subsection{Distillation of datasets and models}

\looseness=-1
\myparnodot{Knowledge distillation}~\cite{hinton2014distilling,bucilua2006model} is a mechanism to transfer knowledge from a pretrained ``teacher'' model into a ``student'' one, 
and it usually requires images.
Our approach can be seen as performing {\em image-free} distillation from a generic text-to-image generation model into a specific classification model.
We assume no access to images to distill from and, instead of distilling the visual encoder of the image generation model,
inspired by recent works in NLP~\cite{ma2022prompting},
we prompt a generation model to produce synthetic images and train a classifier with them.

\myparnodot{Dataset distillation}~\cite{cazenavette2022dataset,zhao2021dataset}, 
on the other hand, is a way of compressing a training set of real images into a smaller set of synthetic images such that after training a model on those, it performs as well as if it had been trained on the original set.
However, one needs to tailor the generation process to a specific task, whereas in our case, we sample images from a task-agnostic generator.

\looseness=-1
\myparnodot{Reconstructing images from model activations} can be considered as another form of distillation.
Earlier works reconstruct images from gradient-based features~\cite{vondrick2013hoggles,weinzaepfel2011reconstructing} or CNN activations~\cite{mahendran2015understanding}.
Since then many methods have tried to uncover the training data distribution as it is stored in the weights of 
a model~\cite{yin2020dreaming,chen2019data}.
Instead of trying to recover the training distribution of the teacher image generation model, we use prompting to distill its knowledge 
for a specific image classification task.

%% file: tex/3_preliminaries.tex
\section{Preliminaries}
\label{sec:preliminaries}

In this section, we first define the task we solve, \ie, learning an image classification model when the training set
of real images is
replaced by an image generator, and
training proceeds using only synthetically generated images.
We then briefly describe Stable Diffusion~\cite{rombach2022high}, \ie, the text-to-image generation model we use in this paper.

\mypar{Task formulation}
Our goal is to learn an image classification model given a set of class names $\classset$ and a text-to-image generator $\mathcal{G}$.
This task is a variant of image classification where the fixed-size image training set is replaced by an image generator.
The model we aim
to learn consists of an encoder $\vz = f_\theta(\vx)$ that maps an \emph{image} $\vx$ into a vector representation $\vz \in \mathbb{R}^d$, and a classifier $\vy = q(\vz)$ that outputs a
distribution $\vy$ over the $\nclasses$ classes $c_i \in \classset$, where $i =\{1,..,\nclasses\}$.
We follow the common supervised learning setting~\cite{russakovsky2015ilsvrc, krizhevsky2012alexnet} and, unless otherwise stated, learn the encoder parameters $\theta$ together with the classifier $q$ for the task.
This model (encoder and classifier) is evaluated on the initial classification task, by applying it to real images (Sec.~\ref{sec:exp_imagenet} and Sec.~\ref{sec:exp_generalization}).
We also evaluate
the visual encoder in the context of several transfer learning tasks (Sec.\ref{sec:exp_transfer}).

\mypar{Text-to-image with Stable Diffusion}
We use the recent Stable Diffusion model~\cite{rombach2022high} (SD) as text-to-image generator $\mathcal{G}$.
SD is a denoising diffusion model~\cite{ho2020denoising} built around the idea of \textit{latent diffusion}.
The diffusion process is run on a compressed latent space for efficiency.
An image encoder/decoder is used to interface the latent diffusion model with the pixel space.
The generation process can be conditioned in many ways, \eg, with text for text-to-image generation, or an image latent vector for image manipulation.

The text-to-image SD model consists of three main components:
i) an autoencoder whose visual encoder outputs a structured latent representation that is fed as input to the forward diffusion process and whose decoder is then used to convert the latent vectors back to pixels,
ii) a denoising U-Net that
runs the diffusion process, and
iii) a text encoder, \ie, similar to the one used by CLIP~\cite{radford2021clip}.

The text-to-image generation process takes a textual prompt $p$ as input
and generates an image $\vx \in \mathbb{R}^{W\times H\times 3}$. Let $g(p)$ denote the generation function of model $\mathcal{G}$. Image $\vx$ is then given by $\vx = g(p)$. In practice, the prompt $p$ is first encoded via the text encoder and the text embedding is used as a conditioning vector for the latent diffusion process that runs for a number of {steps}.
The latent representation is then provided to the decoder, which outputs the image $\vx$.

\looseness=-1
There are two important parameters that control the quality and speed of text-conditioned diffusion; the number of diffusion steps and the coefficient that weights the textual conditioning vector.
The former is linearly related to extraction time, while
the latter provides an excellent way of controlling the visual diversity of generated images. The default values are 50 steps and guidance scale equal to 7.5.

\mypar{Link to distillation}
Since the generator is a model that internally encodes visual information, the image classification model we learn is essentially derived from $\mathcal{G}$.
Under this formulation, and as discussed in~\Cref{sec:related}, one can also see this task as text-guided, image-free knowledge distillation.
Here we distill knowledge from a model of a very different nature, \ie, a text-to-image generation model, to a purely visual encoder, for solving a specific task.

%% file: tex/4_method.tex
\section{Generating synthetic ImageNet clones}
\label{sec:imagenet_sd}

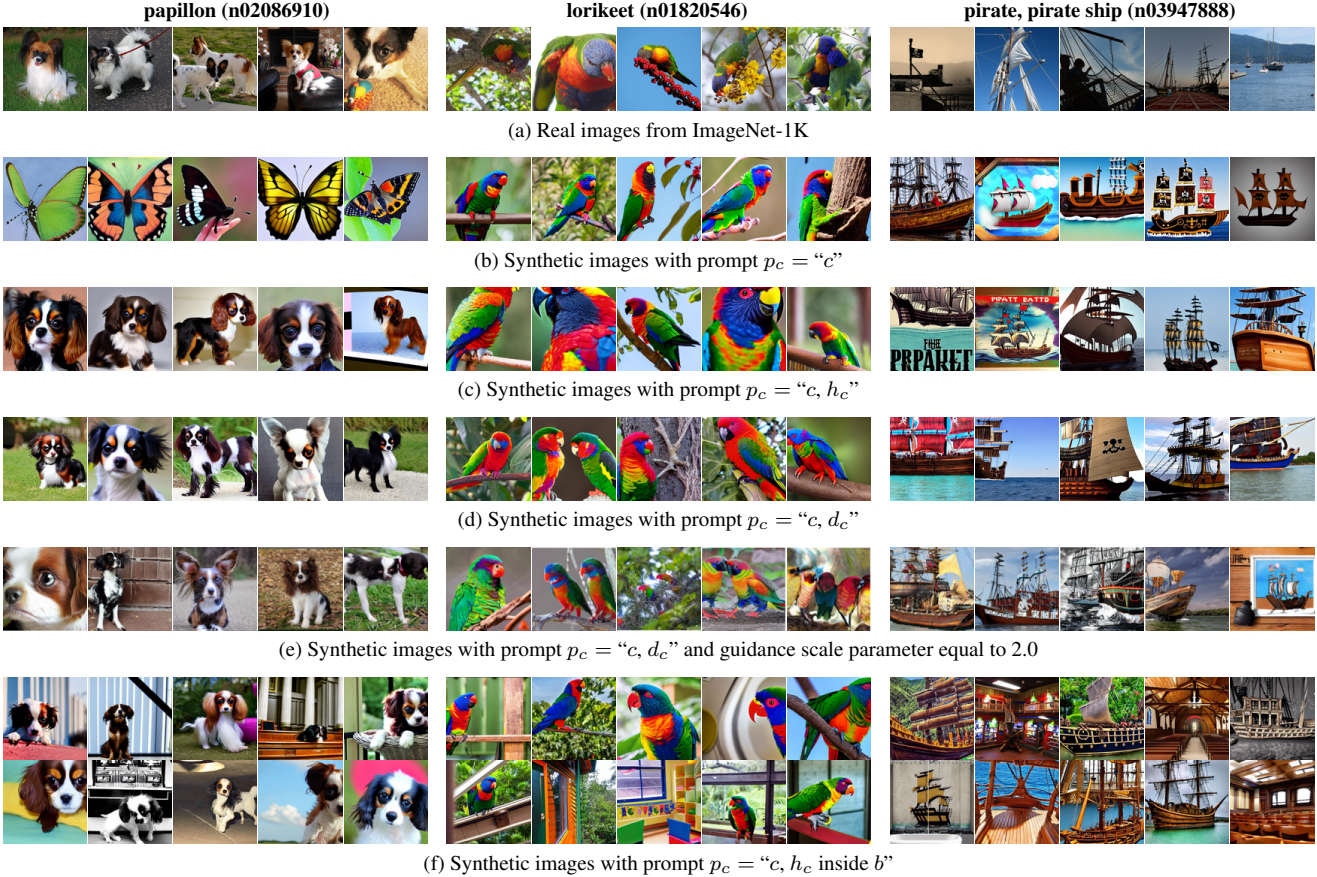
\begin{figure*}[t!]
    \input{fig/qualitative}
    \vspace{-0.08cm}
    \caption{
        \textbf{Qualitative results.}
        (a) Real ImageNet images.
        (b)-(g) Synthetic ImageNet-SD images generated with different prompts.
        Despite high photo-realistic quality, some issues are noticeable for (b) such as i) semantic errors \eg, for the class ``papillon'', ii) lack of diversity, and iii) distribution shifts \eg, towards cartoons for the ``pirate" class.
        Such issues are addressed with more expressive prompts in (c)-(g).}
    \vspace{-0.08cm}
    \label{fig:qualitative}
    \vspace{-4pt}
\end{figure*}

For our study, we create clones of the ImageNet~\cite{deng2009imagenet} dataset by synthesizing images depicting the classes it contains.
We refer to all synthetic datasets of ImageNet classes that are created using Stable Diffusion as \textbf{ImageNet-SD}.
\Cref{sec:baseline} describes different ways of creating ImageNet-SD datasets starting from simply using the class name as the prompt.
We then present generic, class-agnostic ways for tackling issues that arise with respect to semantics and diversity in~\Cref{sec:semantics,sec:diversity}, respectively.
We present a few sample qualitative results in~\Cref{fig:qualitative}, with a more extensive set in the supplementary material.

\subsection{Generating datasets using class names}
\label{sec:baseline}

In the absence of a training set of real images, we use the generator $\mathcal{G}$ presented in the previous section to synthesize images for each class in the set $\classset$.
To do so, we need to provide the generator with at least one prompt per class.
When used as an input, this class-conditioned prompt $p_c$ triggers the generation of a synthetic image $\vx_c = g(p_c)$ from class $c$.
The simplest prompt one could think of is the class name \ie, $p_c = $~``$c$''.
Although CLIP~\cite{radford2021clip} uses $p_c = $~``a photo of a $c$'' for their zero-shot experiments,
using only the class name gives better results in our case.

Each class in ImageNet is associated with one or more \textit{synsets}, \ie, entities, in the WordNet~\cite{miller1995wordnet} graph.
We use the synset lemmas corresponding to each class as class-name prompt ``$c$'', comma-separated if
more than one.
\Cref{fig:im100_c} shows random examples of images generated with such prompts.
At first glance, one can appreciate the ability of the generator to create photo-realistic images given only a class name.
In~\cref{sec:experiments}, we show that one can already obtain surprisingly good image classification results by simply training a model with this synthetic dataset.

Upon close inspection of the generated images, however, some issues become apparent:
a)~\textbf{semantic errors}: Images generated for some classes may capture the wrong semantics (\eg, see the ``papillon'' class in~\Cref{fig:im100_c}),
b)~\textbf{lack of diversity}: Generated images tend to look alike (an issue more apparent in the supplementary material,
and
c)~\textbf{visual domain issues}: some classes tend to shift away from natural images towards sketches or art (\eg, the ``pirate ship'' class in~\Cref{fig:im100_c}).
We discuss and address these issues
in the following.

\subsection{Addressing issues with semantics and domain}
\label{sec:semantics}

As mentioned
earlier, by comparing the (real) images from ImageNet with the synthetic ones generated using only synset names as prompts, we observe that for some classes their semantics do not match.
This is due to polysemy,
\ie, multiple semantic meanings or physical instantiations of the class names we used as prompt.
We show one such case in the left-most column of~\Cref{fig:im100_c}: the ``papillon'' images correspond to butterfly for our generated dataset, while the ImageNet synset contains images of the dog breed of the same name (see~\Cref{fig:im100_real}).

To reduce this semantic ambiguity,
we leverage once again the fact that
class names correspond to WordNet~\cite{miller1995wordnet} synsets.
We augment the prompt for class name $c$ with two additional
elements provided by WordNet:
a) The \textit{hypernyms} $h_c$ of the synset as defined by the WordNet graph, \ie, the class name(s) of the parent node(s) of this class in the graph; and b) the \textit{definition} $d_c$ of the synset, \ie, a sentence-length description of the semantics of each synset. In both cases, we append this information to the prompt, which becomes $p_c = $~``$c$, $h_c$'' and $p_c = $~``$c$, $d_c$'' for hypernyms and definition, respectively.

Qualitatively, we observed that issues regarding the semantics of the most problematic classes are fixed, and so are, to some extent, issues related to visual domain mismatch.
These are
also visible in~\Cref{fig:im100_p,fig:im100_d}: appending the hypernym ($h_c=$~``toy spaniel'') or the description ($d_c=$~``small slender toy spaniel with erect ears and a black-spotted brown to white coat'') of the class ``papillon'' in the prompt
produces images with the dog breed as the main subject. Appending the hypernym ($h_c=$~``ship'') or the description ($d_c=$~``a ship that is manned by pirates'') of the class ``pirate ship'' results in more natural-looking images rather than illustrations, reducing the domain shift.

\subsection{Increasing the diversity of generated images}
\label{sec:diversity}

Generating images using more expressive prompts, \eg, by appending class hypernym or definition, not only reduces semantic errors, but also increases the visual diversity of the output images. This is visible, for example, in the ``lorikeet'' and ``pirate ship'' classes in~\Cref{fig:im100_p,fig:im100_d} when compared to~\Cref{fig:im100_c}: the pose and viewpoints are slighly more diverse.
However, images still tend to display
the class instance centered and
in a prominent position.
The real ImageNet images  feature significantly more diversity, several different settings and backgrounds, and, in several cases, multiple instances of the same class (\eg, see~\Cref{fig:im100_real}).

Although class-specific prompt engineering is an appealing option,
in this study we chose to remain generic, and to increase diversity
in class-agnostic ways.

\mypar{Reducing reliance on the textual prompt}
The text-conditioned generation process of Stable Diffusion uses
classifier-free diffusion guidance~\cite{ho2021classifierfree}
which jointly trains both the conditional and unconditional diffusion models, and combines their
estimates, resulting in
a trade-off between sample quality and diversity. This trade-off is controlled by the guidance scale parameter, that has in practice been shown to produce high-quality images in the range of 6-9 (the default value is 7.5). Although visually detailed (see \Cref{fig:im100_c,fig:im100_p,fig:im100_d}), the resulting images lack diversity. We therefore experiment with reducing the guidance scale. Despite a small degradation in the visual quality of the generated images, setting the scale to 2.0 results in more diverse sets of images as shown in~\Cref{fig:im100_scaletwo}.

\mypar{Diversifying the background} We assume that class $c$ can be seen ``inside'' a scene or background.
To remain class-agnostic, we
use {all} the scene classes from the Places dataset~\cite{zhou2017places} as background for every class.
We generate images for every possible combination of a class $c$ and a scene $b \in \mathcal{B}$ from the set $\mathcal{B}$ of 365 scenes in Places.
We found that ``$c$ inside $b$'' generally
produces the best-looking results among a few prepositions we tried. However, we found that semantic and domain errors that arise from generating only using class name remained after specifying a background. We therefore build on top of the second simplest, but more semantically correct prompt variant, and use $p_c =$~``$c$, $h_c$ inside $b$''
to generate images in diverse scenes and backgrounds.
Although we do not consider this in our study, selecting backgrounds tailored for each class, \eg, by matching class names to scenes using features from a text encoder, seems like a promising future direction.

\mypar{Label noise and visual realism}
Quite a few generated images, especially those with low guidance scale parameters or  with random backgrounds (\eg, see~\Cref{fig:im100_bg,fig:im100_scaletwo})
are not realistic,
for example, the right-most image in the first column of~\Cref{fig:im100_scaletwo}.
When the prompt mentions a background, some images miss the foreground object completely (\eg, see the bottom row in the middle column of~\Cref{fig:im100_bg}) or contain
impossible combinations of objects and scenes. Yet, we see such noisy or unrealistic synthetic images as a way of adding stochasticity during the training process, similar to what strong non-realistic data augmentation achieves~\cite{geiping2023how,xu2021robust}.
In fact, it was recently shown~\cite{geiping2023how} that diverse data augmentations, even when inconsistent with the data distribution, can be valuable (even more than additional training data) for out-of-distribution scenarios.
Our experimental validation
corroborates this claim.

%% file: fig/qualitative.tex
\centering

\resizebox{\linewidth}{!}{
    \begin{tabular}{ccc}
    \quad\quad\quad\quad\quad\quad\textbf{papillon (n02086910)} \hspace{100pt} &
    \textbf{lorikeet (n01820546)} \hspace{100pt} &
    \hspace{-10pt}\textbf{pirate, pirate ship (n03947888)}\quad\quad\quad\quad \\
    \end{tabular}
}

\begin{subfigure}[t]{\linewidth}
    \centering
    \resizebox{\linewidth}{!}{
        \begin{tabular}{ccccc}
                \includegraphics[width=\textwidth]{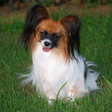} &
                \includegraphics[width=\textwidth]{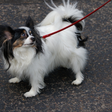} &
                \includegraphics[width=\textwidth]{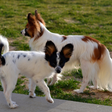} &
                \includegraphics[width=\textwidth]{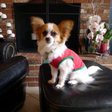} &
                \includegraphics[width=\textwidth]{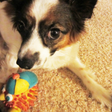}
        \end{tabular} \hspace{100pt}
        \begin{tabular}{ccccc}
                \includegraphics[width=\textwidth]{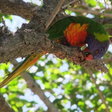} &
                \includegraphics[width=\textwidth]{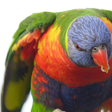} &
                \includegraphics[width=\textwidth]{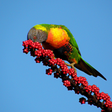} &
                \includegraphics[width=\textwidth]{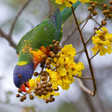} &
                \includegraphics[width=\textwidth]{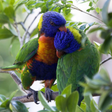}
        \end{tabular} \hspace{100pt}
        \begin{tabular}{ccccc}
                \includegraphics[width=\textwidth]{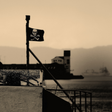} &
                \includegraphics[width=\textwidth]{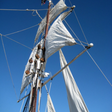} &
                \includegraphics[width=\textwidth]{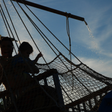} &
                \includegraphics[width=\textwidth]{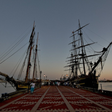} &
                \includegraphics[width=\textwidth]{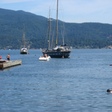}
        \end{tabular}
    }
    \caption{Real images from \imnetlong}
    \label{fig:im100_real}
\end{subfigure}

\begin{subfigure}[t]{\linewidth}
    \centering
    \resizebox{\linewidth}{!}{
        \begin{tabular}{ccccc}
                \includegraphics[width=\textwidth]{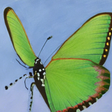} &
                \includegraphics[width=\textwidth]{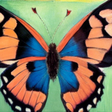} &
                \includegraphics[width=\textwidth]{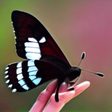} &
                \includegraphics[width=\textwidth]{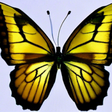} &
                \includegraphics[width=\textwidth]{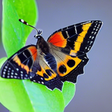}
        \end{tabular} \hspace{100pt}
        \begin{tabular}{ccccc}
                \includegraphics[width=\textwidth]{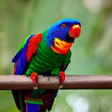} &
                \includegraphics[width=\textwidth]{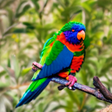} &
                \includegraphics[width=\textwidth]{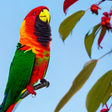} &
                \includegraphics[width=\textwidth]{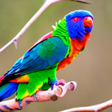} &
                \includegraphics[width=\textwidth]{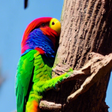}
        \end{tabular} \hspace{100pt}
        \begin{tabular}{ccccc}
                \includegraphics[width=\textwidth]{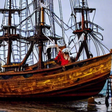} &
                \includegraphics[width=\textwidth]{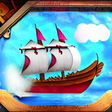} &
                \includegraphics[width=\textwidth]{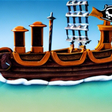} &
                \includegraphics[width=\textwidth]{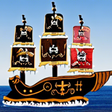} &
                \includegraphics[width=\textwidth]{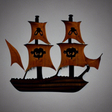}
        \end{tabular}
    }
    \caption{Synthetic images with prompt $p_c = $~``$c$''}
    \label{fig:im100_c}
\end{subfigure}

\begin{subfigure}[t]{\linewidth}
    \centering
    \resizebox{\linewidth}{!}{
        \begin{tabular}{ccccc}
                \includegraphics[width=\textwidth]{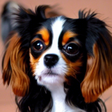} &
                \includegraphics[width=\textwidth]{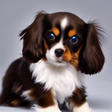} &
                \includegraphics[width=\textwidth]{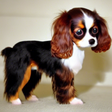} &
                \includegraphics[width=\textwidth]{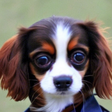} &
                \includegraphics[width=\textwidth]{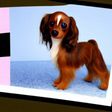}
        \end{tabular} \hspace{100pt}
        \begin{tabular}{ccccc}
                \includegraphics[width=\textwidth]{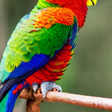} &
                \includegraphics[width=\textwidth]{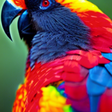} &
                \includegraphics[width=\textwidth]{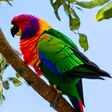} &
                \includegraphics[width=\textwidth]{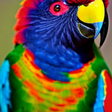} &
                \includegraphics[width=\textwidth]{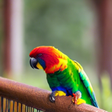}
        \end{tabular} \hspace{100pt}
        \begin{tabular}{ccccc}
                \includegraphics[width=\textwidth]{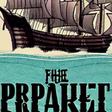} &
                \includegraphics[width=\textwidth]{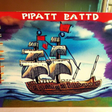} &
                \includegraphics[width=\textwidth]{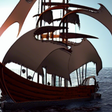} &
                \includegraphics[width=\textwidth]{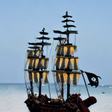} &
                \includegraphics[width=\textwidth]{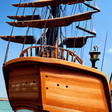}
        \end{tabular}
    }

    \caption{Synthetic images with prompt $p_c = $~``$c$, $h_c$''}
    \label{fig:im100_p}
\end{subfigure}
\begin{subfigure}[t]{\linewidth}
    \centering
    \resizebox{\linewidth}{!}{
        \begin{tabular}{ccccc}
                \includegraphics[width=\textwidth]{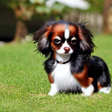} &
                \includegraphics[width=\textwidth]{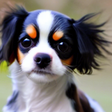} &
                \includegraphics[width=\textwidth]{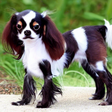} &
                \includegraphics[width=\textwidth]{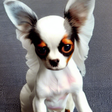} &
                \includegraphics[width=\textwidth]{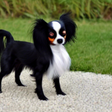}
        \end{tabular} \hspace{100pt}
        \begin{tabular}{ccccc}
                \includegraphics[width=\textwidth]{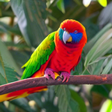} &
                \includegraphics[width=\textwidth]{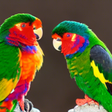} &
                \includegraphics[width=\textwidth]{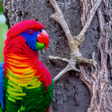} &
                \includegraphics[width=\textwidth]{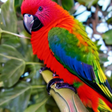} &
                \includegraphics[width=\textwidth]{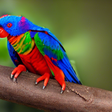}
        \end{tabular} \hspace{100pt}
        \begin{tabular}{ccccc}
                \includegraphics[width=\textwidth]{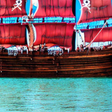} &
                \includegraphics[width=\textwidth]{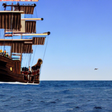} &
                \includegraphics[width=\textwidth]{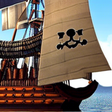} &
                \includegraphics[width=\textwidth]{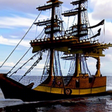} &
                \includegraphics[width=\textwidth]{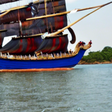}
        \end{tabular}
    }
    \caption{Synthetic images with prompt $p_c = $~``$c$, $d_c$''}
    \label{fig:im100_d}
\end{subfigure}
\begin{subfigure}[t]{\linewidth}
    \centering
    \resizebox{\linewidth}{!}{
        \begin{tabular}{ccccc}
                \includegraphics[width=\textwidth]{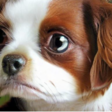} &
                \includegraphics[width=\textwidth]{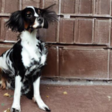} &
                \includegraphics[width=\textwidth]{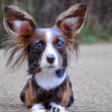} &
                \includegraphics[width=\textwidth]{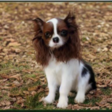} &
                \includegraphics[width=\textwidth]{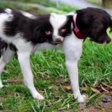}
        \end{tabular} \hspace{100pt}
        \begin{tabular}{ccccc}
                \includegraphics[width=\textwidth]{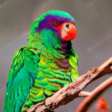} &
                \includegraphics[width=\textwidth]{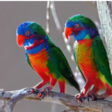} &
                \includegraphics[width=\textwidth]{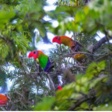} &
                \includegraphics[width=\textwidth]{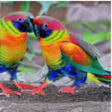} &
                \includegraphics[width=\textwidth]{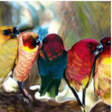}
        \end{tabular} \hspace{100pt}
        \begin{tabular}{ccccc}
                \includegraphics[width=\textwidth]{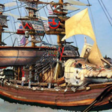} &
                \includegraphics[width=\textwidth]{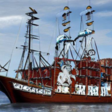} &
                \includegraphics[width=\textwidth]{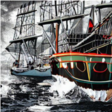} &
                \includegraphics[width=\textwidth]{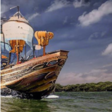} &
                \includegraphics[width=\textwidth]{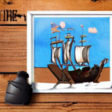}
        \end{tabular}
    }
    \caption{Synthetic images with prompt $p_c = $~``$c$, $d_c$'' and guidance scale parameter equal to 2.0}
    \label{fig:im100_scaletwo}
\end{subfigure}
\begin{subfigure}[t]{\linewidth}
    \centering
    \resizebox{\linewidth}{!}{
        \begin{tabular}{ccccc}
                \includegraphics[width=\textwidth]{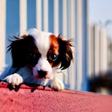} &
                \includegraphics[width=\textwidth]{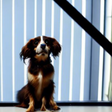} &
                \includegraphics[width=\textwidth]{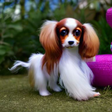} &
                \includegraphics[width=\textwidth]{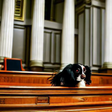} &
                \includegraphics[width=\textwidth]{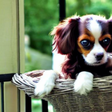} \\
                \includegraphics[width=\textwidth]{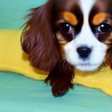} &
                \includegraphics[width=\textwidth]{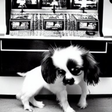} &
                \includegraphics[width=\textwidth]{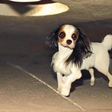} &
                \includegraphics[width=\textwidth]{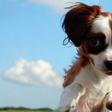} &
                \includegraphics[width=\textwidth]{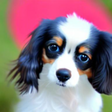}

        \end{tabular} \hspace{100pt}
        \begin{tabular}{ccccc}
                \includegraphics[width=\textwidth]{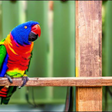} &
                \includegraphics[width=\textwidth]{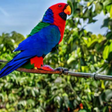} &
                \includegraphics[width=\textwidth]{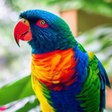} &
                \includegraphics[width=\textwidth]{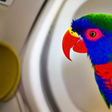} &
                \includegraphics[width=\textwidth]{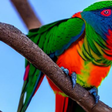} \\
                \includegraphics[width=\textwidth]{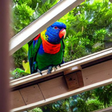} &
                \includegraphics[width=\textwidth]{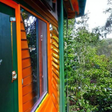} &
                \includegraphics[width=\textwidth]{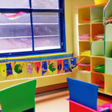} &
                \includegraphics[width=\textwidth]{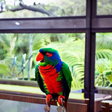} &
                \includegraphics[width=\textwidth]{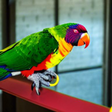}
        \end{tabular} \hspace{100pt}
        \begin{tabular}{ccccc}
                \includegraphics[width=\textwidth]{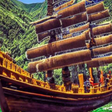} &
                \includegraphics[width=\textwidth]{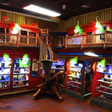} &
                \includegraphics[width=\textwidth]{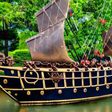} &
                \includegraphics[width=\textwidth]{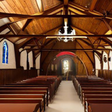} &
                \includegraphics[width=\textwidth]{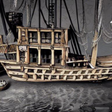} \\
                \includegraphics[width=\textwidth]{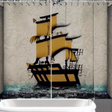} &
                \includegraphics[width=\textwidth]{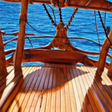} &
                \includegraphics[width=\textwidth]{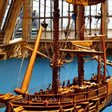} &
                \includegraphics[width=\textwidth]{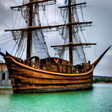} &
                \includegraphics[width=\textwidth]{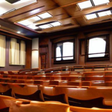}
        \end{tabular}
    }
    \caption{Synthetic images with prompt $p_c = $~``$c$, $h_c$ inside $b$''}
    \label{fig:im100_bg}
\end{subfigure}

%% file: tex/5_experiments.tex
\section{Experiments}\label{sec:experiments}

\begin{figure}[t]
    \centering
    \begin{subfigure}{.99\linewidth}
        \includegraphics[width=\linewidth]{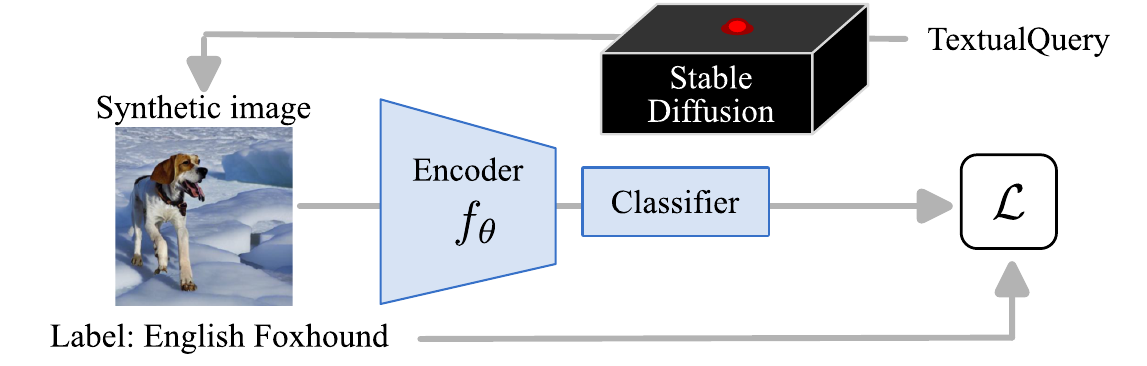}
        \caption{Training a model on synthetic images.}
        \label{fig:protocol_train}
    \end{subfigure}
    \begin{subfigure}{.99\linewidth}
        \vspace{1pt}
        \includegraphics[width=\linewidth]{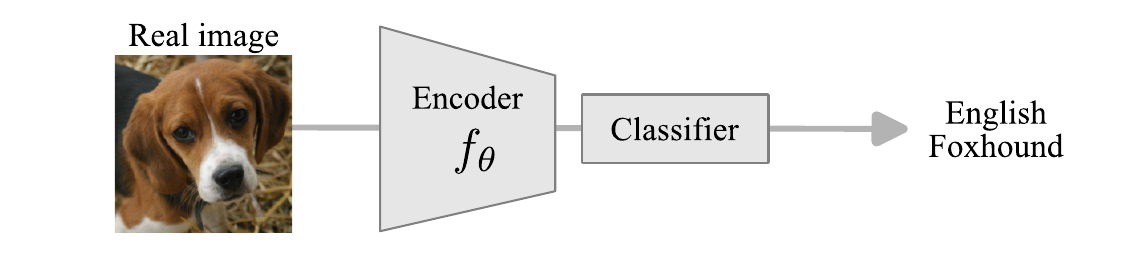}
        \caption{Testing the frozen model on real images.}
        \label{fig:protocol_test}
    \end{subfigure}
    \caption{
    {\bf Overview of our experimental protocol.}
    During training, the model has access to synthetic images generated by the Stable Diffusion model, provided with a set of prompts per class.
    During evaluation, real images are classified by the frozen model.
    }
    \label{fig:protocol}
\end{figure}

In this section we analyze the performance of image classification models learned using the different synthetic datasets constructed as described in~\Cref{sec:imagenet_sd}.
Due to the size of \imnetlong (roughly 1.3 million images), we perform most of our study on the smaller \imnethlong~\cite{tian2019cmc} dataset.
This allows us to run multiple flavours of each synthetic dataset and to measure the impact of several design choices.
Because \imnethlong is a randomly chosen subset of \imnetlong, spanning over 100 classes and 126,689 images, it preserves some important characteristics of \imnetlong
such as its fine-grained nature.

We denote synthetic datasets for the two ImageNet subsets as \textbf{\imnethsdlong}~(\imnethsd) and \textbf{ImageNet-1K-SD} (\imnetsd), respectively.

\input{tab/imagenet.tex}

\looseness=-1
\mypar{Experimental protocol} We follow the protocol illustrated in~\Cref{fig:protocol}.
The generator $\mathcal{G}$ is the Stable Diffusion~\cite{rombach2022high} v1.4 model,\footnote{\scriptsize\url{https://huggingface.co/CompVis/stable-diffusion-v1-4}}
trained on the LAION2B-en dataset~\cite{schuhmann2022laion} and
fine-tuned on a smaller subset
filtered by an
aesthetics classifier.
During training, the generator is used to synthesize images for each class,
which are then used for training the parameters of the encoder and the classifier.
Unless otherwise stated, we create datasets of the {exact same size} as their real-image counterparts, \ie, we generate the exact same number of images for every class as in the corresponding real dataset, maintaining any class imbalance.

\looseness=-1
We evaluate all the models on {real images}.
When evaluating their performance over the ImageNet classes, we use both the
encoder and the classifier learned
during training to predict labels of real images for the 5 ImageNet datasets (\Cref{sec:exp_imagenet,sec:exp_generalization}).
For transfer learning (\Cref{sec:exp_transfer}), we use the pretrained encoder as a feature extractor, and learn a separate linear classifier on each of the {15} transfer datasets.

All our experiments use ResNet50~\cite{he2016resnet} as the encoder $f_\theta$.
Unless otherwise stated, we use 50 diffusion steps.
We provide ablations for the diffusion steps and guidance scale as well as more implementation details in the supplementary material.
We use multi-crop data augmentation~\cite{caron2021dino},
as it results in large performance gains for the models trained on ImageNet-SD (see supplementary for more details).
Indeed, strong transformations have been shown to improve domain generalization~\cite{volpi2021cvpr}, and to reduce the  sim-to-real gap.

\subsection{Results on ImageNet datasets}
\label{sec:exp_imagenet}

\mypar{Evaluating different prompts on \imnethlong}
\Cref{tab:imagenet} compares the performance of models trained using variants of ImageNet-100-SD created with the different prompts presented in~\Cref{sec:imagenet_sd}, for two different guidance scale values: $7.5$ and $2$.
From
the results for ImageNet-val and ImageNet-v2 (four left-most columns), we make the following observations:
\textbf{(a)} Simply using the class name as a prompt and the default guidance scale (row 2), one can synthesize images and learn a visual encoder \textit{from scratch} that already achieves \emph{more than $70\%$ Top-5 accuracy} ($43\%$ Top-1 accuracy) on \imnethlong{}, a challenging 100-way classification task with many fine-grained classes.
\textbf{(b)} Adding the hypernym or the definition from WordNet as part of the prompt (rows 3, 4) addresses some of the semantic and domain issues and translates into performance gains.
\textbf{(c)} Generating objects on diverse backgrounds (row 5), even in a simple and class-agnostic way,
gives the best results for the default guidance scale, reaching over $50\%$ Top-1 and $76\%$ Top-5 accuracy on \imnethlong.
\textbf{(d)} Using a lower guidance scale value (2) leads to more diverse image sets (as discussed in~\Cref{sec:diversity}) and translates into the best overall performance on \imnethlong.
\textbf{(e)} The exact formulation of the prompt has less impact
when
lowering the guidance scale; all the four prompt variants lead to similar performance as we see from rows 6-9.

\mypar{Scaling the number of synthetic images}
Unlike real datasets that are capped in the number of images they contain, ImageNet-SD has theoretically no size upper bound as one can generate images on demand.
We therefore generated datasets which are $10\times$, $20\times$ and $50\times$ larger than ImageNet-100,
using
prompt $p_c=$ ``$c$, $d_c$'' (the best variant in~\Cref{tab:imagenet}, row 8) for the classes of \imnethlong.
From the last three rows of the top section in~\Cref{tab:imagenet}, we see that this brings gains of up to $8.5\%$ in Top-1 accuracy on \imnethlong, with our best model reaching $73.3\%$ Top-1 (and $91.7\%$ Top-5) accuracy.
The gains are even more prominent for transfer learning, as we discuss in~\Cref{sec:exp_transfer}.

\mypar{Results on \imnetlong}
In the bottom
part of~\Cref{tab:imagenet} we report results on the very challenging 1000-way classification task of \imnetlong (IN-Val) that contains many fine-grained categories of mushrooms, birds and dogs~\cite{huh2016what}.
We see that the model trained on our synthetic ImageNet-1K-SD dataset using the prompt composed of the class name and description ($p_c=$ ``$c$, $d_c$'') and using guidance scale 2 reaches
$42.9\%$ Top-1 and $70.3\%$ Top-5 accuracy on the \imnetlong validation set.
Although significantly lower than the results achieved by a model trained on the 1.3 million real images of ImageNet, we see that the synthetic dataset is able to at least partially capture the subtle clues needed to differentiate fine-grained classes.
Similar observations can be made on ImageNet-v2~\cite{recht2019imagenet} (IN-v2).

\input{tab/transfer.tex}

\subsection{Resilience to domain shifts}
\label{sec:exp_generalization}

\looseness=-1
We investigate the performance of our models on three challenging evaluation sets for ImageNet-1K classes:
ImageNet-Sketch~\cite{wang2019learning} (IN-Sketch), ImageNet-R~\cite{hendrycks2021many} (IN-R) and ImageNet-A~\cite{hendrycks2021natural} (IN-A).
These datasets contain out-of-distribution images and their goal is to test resilience to domain shifts and adversarial images.
Results are reported in the right-most columns of~\Cref{tab:imagenet}.

For ImageNet-100, we see from the top part of the table that a number of ImageNet-100-SD models \textit{outperform} the model trained on
real images
for ImageNet-Sketch and ImageNet-R. The best Imagenet-100-SD model, \ie the one trained with $50\times$ images, further rivals the baseline on ImageNet-A.

When it comes to a much harder classification task like the 1000 classes of ImageNet-1K,
we see from the
lower part of~\Cref{tab:imagenet} that the same trend does not really hold.
The ImageNet-1K-SD model trained on synthetic data lags behind in all cases when compared to the two
models~\cite{pytorch,wightman2021rsb} that are trained on the \imnetlong training set.

\subsection{Transfer learning}\label{sec:exp_transfer}

In previous evaluations, we used pretrained models as a whole, \ie, encoders together with classifiers, all trained on synthetic ImageNet datasets, and we directly applied those to predict the label of the (real) test images on the training classes.
Here, we use a slightly different protocol.
We evaluate the quality of the representations learned by our encoders alone, by using them as feature extractors and training linear logistic regression classifiers from scratch on top as done in transfer learning~\cite{kornblith2019transfer,sariyildiz2021cog}.

We report results on 15 transfer datasets: \textbf{(a)} eight common small-scale datasets (Aircraft~\cite{maji2013aircraft},
Cars196~\cite{krause2013cars},
DTD~\cite{cimpoi2014texture},
EuroSAT~\cite{helber2019eurosat},
Flowers~\cite{nilsback2008flowers},
Pets~\cite{parkhi2012pets},
Food101~\cite{bossard2014food101},
SUN397~\cite{xiao2010sun}), \textbf{(b)} two long-tail datasets (iNat2018~\cite{van2018inaturalist} and iNat2019~\cite{van2018inaturalist}), and \textbf{(c)} the five datasets (``levels'') of the CoG benchmark~\cite{sariyildiz2021cog}.
We report Top-1 accuracy on the (real) test set of
the small-scale and long-tail datasets in~\Cref{tab:transfer}.
In~\Cref{fig:teaser} and the supplementary, we present results on the CoG benchmark.
We compare \imnethsdlong{} and \imnetsdlong{} visual encoders obtained with some of our best prompts to baselines trained on \imnethlong and \imnetlong{}. What we observe is quite striking:
On average, representations learned on purely synthetic images exhibit \emph{generalization performance comparable to representations trained on thousands or millions of real images}.
This suggests that synthetic images can be used to pretrain strong general-purpose visual encoders.

Following this transfer learning protocol,
our best model achieves \emph{70.4\%} Top-1 accuracy on ImageNet-1K (evaluation as part of the CoG benchmark, detailed in the supplementary material),
significantly closing the gap to models trained on real data.
This protocol differs from the one presented in \Cref{sec:exp_imagenet} as it uses real images to train a linear classifier on top of the feature extractor trained only on synthetic images, hence results are not comparable with \Cref{tab:imagenet}.

\begin{figure}[t!]
    \centering
    \vspace{-5pt}
    \input{fig/scaling_plot_transfer_main}
    \caption{{\bf Scaling the number of training images.} Average top-1 accuracy on 10 transfer datasets when training on \imnethlong using $(1/10)$-th to 50$\times$ images (relative to the real dataset size).}
    \label{fig:scaling}
\end{figure}
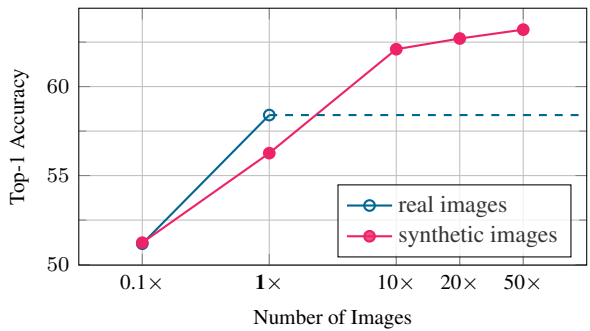

\mypar{Scaling the number of synthetic images for transfer} \Cref{fig:scaling} reports transfer performance on the 10
datasets of~\Cref{tab:transfer}, when varying the size of the training set. We see that generating 10$\times$ more images allows
the \imnethsdlong model to outperform the model trained on real images, and the gains increase as we generate up to $50\times$.

%% file: tab/imagenet.tex
{
\setlength{\tabcolsep}{2pt}
\clearrow
\begin{table*}[t]
    \centering
    \adjustbox{max width=\linewidth}{
    \begin{tabular}{>{\rowmac}l>{\rowmac}c>{\rowmac}c>{\rowmac}l  >{\rowmac}c>{\rowmac}c>{\rowmac}c>{\rowmac}c>{\rowmac}c>{\rowmac}c>{\rowmac}c>{\rowmac}c>{\rowmac}c>{\rowmac}c <{\clearrow}}
        \toprule
        \multicolumn{2}{l}{Training Dataset}
            & \multirow{2}{0.85cm}{\centering Scale}
            & \multirow{2}{5cm}{\quad \quad Prompt ($p_c$) / \textit{Model}}
            & \multicolumn{2}{c}{IN-Val}
            & \multicolumn{2}{c}{IN-v2}
            & \multicolumn{2}{c}{IN-Sketch}
            & \multicolumn{2}{c}{IN-R$^*$}
            & \multicolumn{2}{c}{IN-A$^*$} \\
        &  R. Size & & & Top-1 & Top-5 & Top-1 & Top-5 & Top-1 & Top-5 & Top-1 & Top-5 & Top-1 & Top-5 \\
        \specialrule{2pt}{3pt}{5pt}
        \setrow{\itshape} \imnethlong & & -- & \myrowno{1} Baseline & 87.4 & 96.8 & 82.5 & 95.1 & 39.1 & 58.9 & 58.4 & 79.1 & 25.6 & 68.7 \clearrow \\
        \midrule
        \multirow{10}{*}{\imnethsdlong} & & \multirow{4}{*}{7.5} & \myrowno{2} $p_c=$ ``$c$''                                        & 43.1 & 70.7 & 45.4 & 70.7 & 29.9 & 53.5 & 51.7 & 75.3 &  8.8 & 38.4 \\
        &                                 &                      & \myrowno{3} $p_c=$ ``$c$, $h_c$''                                 & 46.9 & 73.4 & 47.3 & 73.7 & 25.9 & 50.4 & 46.3 & 75.3 & 11.5 & 42.2 \\
        &                                 &                      & \myrowno{4} $p_c=$ ``$c$, $d_c$''                                 & 47.9 & 74.2 & 49.1 & 74.9 & 24.7 & 49.2 & 41.2 & 71.5 & 12.2 & 38.5 \\
        &                                 &                      & \myrowno{5} $p_c=$ ``$c$, $h_c$ inside $b$''                      & 51.5 & 76.8 & 51.2 & 77.4 & 27.9 & 52.5 & 54.0 & {\better{81.8}} & 14.1 & 48.4 \\
        \cmidrule{3-14}
        & & \multirow{4}{*}{2.0}          & \myrowno{6} $p_c=$ ``$c$''                   & 63.5 & 86.9 & 62.7 & 86.7 & \better{41.8} & \better{\bf 67.6} & \better{\bf 64.2} & \better{83.9} & 13.7 & 45.1 \\
        & &                               & \myrowno{7} $p_c=$ ``$c$, $h_c$''            & 63.4 & 87.1 & 63.5 & 86.5 & \better{39.2} & \better{66.7} & \better{61.9} & \better{85.1} & 14.9 & 49.1 \\
        & &                               & \myrowno{8} $p_c=$ ``$c$, $d_c$''            & 64.8 & 86.9 & 65.0 & 87.3 & 33.8 &          \better{60.5} & 51.4          & 77.5          & 14.0 & 48.8 \\
        & &                               & \myrowno{9} $p_c=$ ``$c$, $h_c$ inside $b$'' & 63.1 & 85.7 & 62.0 & 85.0 & 38.7 &          \better{65.5} & \better{64.0} & \better{\bf 87.2} & {\bf 21.9} & {\bf 63.1}\\
        \cmidrule{3-14}
        & 10$\times$ & \multirow{3}{*}{2.0} & \myrowno{10} $p_c=$ ``$c$, $d_c$'' & 72.4 & 90.8 & 70.2 & 90.2 & \better{40.0} & \better{65.7} & 55.2 & 79.0 & 15.6 & 53.8 \\
        & 20$\times$ &                      & \myrowno{11} $p_c=$ ``$c$, $d_c$'' & 72.4 & 91.4 & 71.4 & 90.7 & 38.4          & \better{63.9} & 56.9 & \better{81.5} & {17.8} & 55.0 \\
        & 50$\times$ &                      & \myrowno{12} $p_c=$ ``$c$, $d_c$'' & {\bf 73.3} & {\bf 91.7} & {\bf 72.3} & {\bf 91.2} & \better{\bf 42.0} & \better{67.0} & \better{59.4} & \better{82.3} & {17.1} & {57.1} \\
        \specialrule{2pt}{3pt}{5pt}
        \setrow{\itshape}
        \multirow{2}{*}{\imnetlong}
        & & -- & \myrowno{13} {PyTorch~\cite{torchvision}} & 76.1 & 92.9 & 71.1 & 90.4 & 24.1 & 41.3 & 36.2 & 52.8 & 0.0 & 14.4 \\
        \setrow{\itshape}
        & & -- & \myrowno{14} {{RSB-A1}~\cite{wightman2021rsb}} & 80.1 & 94.5 & 75.6 & 92.0 & 29.2 & 46.5 & 40.6 & 55.1 & 11.1 & 38.6 \clearrow \\
        \midrule
        \multirow{3}{*}{\imnetsdlong}
        & & 7.5 & \myrowno{15} $p_c=$ ``$c$, $d_c$''            & 26.2 & 51.7 & 26.0 & 51.4 & 9.5 & 22.1 & 15.9 & 32.0 & 2.2 & 10.1 \\
        & & 7.5 & \myrowno{16} $p_c=$ ``$c$, $h_c$ inside $b$'' & 30.1 & 55.6 & 29.8 & 55.3 & 11.9 & 27.1 & 23.5 & 43.1 & 3.4 & 13.2 \\
        & & 2.0 & \myrowno{17} $p_c=$ ``$c$, $d_c$''            & {\bf 42.9} & {\bf 70.3} & {\bf 43.0} & {\bf 70.3} & {\bf 16.6} & {\bf 35.1} & {\bf 26.3} & {\bf 45.3} & {\bf 3.6} & {\bf 15.1} \\
        \bottomrule
    \end{tabular}
    }
    \caption{
        \looseness=-1
        \textbf{Results on ImageNet datasets.}
        Top-1 and Top-5 accuracy on several ImageNet datasets, namely IN-Val
        (the ILSVRC-2012 validation set~\cite{russakovsky2015ilsvrc}),
        IN-v2~\cite{recht2019imagenet}, IN-Sketch~\cite{wang2019learning}, IN-R~\cite{hendrycks2021many} and IN-A~\cite{hendrycks2021natural}.
        In all cases, testing is done on real images.
        For the prompts, $h_c$ ($d_c$) refers to the hypernym (definition) of class $c$ provided by WordNet~\cite{miller1995wordnet}, while $b$ to scene classes from Places 365~\cite{zhou2017places}.
        $^*$IN-R and IN-A only cover a subset of the \imnethlong classes and we compute the reported metrics only on the common classes.
        \better{\colorname} scores denote performance higher than the models trained on real images. \emph{Italics} denote results from models trained using real images.
    }
    \label{tab:imagenet}
    \vspace{-4pt}
\end{table*}
}

%% file: tab/transfer.tex
{
\setlength{\tabcolsep}{2pt}
\begin{table*}[t]
    \adjustbox{max width=\linewidth}{
    \begin{tabular}{@{}lclccccccccccc@{}}
    \toprule
    Training Dataset & 
    Scale & 
    \quad \quad Prompt ($p_c$) / \textit{Model}
    & Aircraft & Cars196 & DTD & EuroSAT & Flowers & Pets & Food101 & SUN397 & iNat18 & iNat19 & Avg.\\
    \specialrule{2pt}{3pt}{5pt}
    -- & -- & \myrowno{1} Random Weights & 11.9 & 3.7 & 17.0 & 73.1 & 26.9 & 11.9 & 13.3 & 7.3 & 0.1 & 1.3 & 16.6 \\
    \specialrule{2pt}{3pt}{5pt}
    \textit{\imnethlong} & -- & \myrowno{2} \textit{Baseline}   & \textit{43.6} & \textit{41.5} & \textit{67.9} & \textit{96.2} & \textit{85.6} & \textit{78.7} & \textit{63.4} & \textit{51.2} & \textit{22.8} & \textit{33.4} & \textit{58.4} \\
    \midrule
    \imnethsdlong & 2.0 & \myrowno{3} $p_c=$ ``$c$, $d_c$'' (50$\times$)       & \better{\bf 47.9} & \better{\bf 44.5} & \better{\bf 74.0} & \better{\bf 96.8} & \better{\bf 89.6} & \better{\bf 83.7} & \better{\bf 68.6} & \better{\bf 57.2} & \better{\bf 29.5} & \better{\bf 40.6} & \better{\bf 63.2} \\
    \specialrule{2pt}{3pt}{5pt}
    \multirow{2}{*}{\em \imnetlong}
      & -- & \myrowno{4} {\em PyTorch~\cite{torchvision}}       & \textit{48.9} & \textit{49.9} & \textit{72.1} & \textit{96.2} & \textit{89.3} & \textit{92.3} & \textit{71.2} & \textit{60.5} & \textit{\bf 35.5} & \textit{41.5} & \textit{65.7} \\
      & -- & \myrowno{5} {\em {RSB-A1}~\cite{wightman2021rsb}}  & \textit{46.8} & \textit{54.4} & \textit{73.8} & \textit{95.8} & \textit{88.6} & \textit{\bf 93.0} & \textit{71.3} & \textit{\bf 63.4} & \textit{34.9} & \textit{43.2} & \textit{66.5} \\
    \midrule
    \multirow{3}{*}{\imnetsdlong}
      & 7.5 & \myrowno{6} $p_c=$ ``$c$, $d_c$''                                       & 48.7 & 49.7 & 71.6 & {96.5} & {90.1} & 81.9 & 66.4 & 55.8 & 28.7 & 40.6 & 63.0 \\
      & 7.5 & \myrowno{7} $p_c=$ ``$c$, $h_c$ inside $b$''                            & {49.6} & 47.4 & 72.1 & 95.9 & 89.3 & 87.2 & 67.7 & 59.5 & 30.8 & 41.4 & 64.1 \\
      & 2.0 & \myrowno{8} $p_c=$ ``$c$, $d_c$''                                       & \better{\bf 55.3} & \better{\bf 57.2} & \better{\bf 75.9} & \better{\bf 96.7} & \better{\bf 92.9} & {88.7} & \better{\bf 73.1} & {62.5} & {35.0} & \better{\bf 46.3} & \better{\bf 68.4} \\
    \bottomrule
    \end{tabular}
    }
    \caption{
        {\bf Top-1 accuracy on ten transfer learning datasets} for encoders trained on real and synthetic images.
        We treat encoders as feature extractors and train linear classifiers on top for each dataset.
        \better{\colorname} scores denote performance higher than the models trained on real images.
        We make the remarkable observation that representations from models trained on synthetic data can
        match the generalization performance of representations from models trained on millions of real images. \emph{Italics} denote results from models trained using real images.
    }
    \vspace{-5pt}
    \label{tab:transfer}
\end{table*}
}

%% file: fig/scaling_plot_transfer_main.tex
\begin{tikzpicture}
    \begin{axis}[
        height=5cm,
        xlabel={Number of Images},
        ylabel={Top-1 Accuracy},
        label style={font=\footnotesize},
        xmin=2,
        xmax=10,
        minor y tick num=1,
        xtick={3,5,7,8,9},
        xticklabels={0.1$\times$,\textbf{1$\times$},10$\times$,20$\times$,50$\times$},
        tick label style={font=\footnotesize},
        legend pos=south east,
        legend style={font=\small},
        ]

        \addplot[real1]
        coordinates {
         (3, 51.17)
         (5, 58.4)
        }; \leg{real images};

         \addplot[sd1]
        coordinates {
         (3, 51.23)
         (5, 56.26)
         (7, 62.1)
         (8, 62.7)
         (9, 63.2)
        }; \leg{synthetic images};

        \addplot[real5, mark size=0pt]
        coordinates {
         (5, 58.4)
         (7, 58.4)
         (8, 58.4)
         (9,  58.4)
         (10,  58.4)
        };

    \end{axis}
\end{tikzpicture}

%% file: tex/6_discussion.tex
\section{Discussion}
\label{sec:discussion}

This section takes a step back and considers some of the implications from the analysis proposed in this paper.

\looseness=-1
\mypar{Applicability beyond ImageNet}
The process we followed to create ImageNet-SD requires minimal assumptions and can be applied to a wider set of classes.
To disambiguate semantics, we only assume access to a short textual description of the class.
This is generally easy to acquire even at a larger scale, \eg, in semi-automatic ways from Wikipedia. 

\looseness=-1
\mypar{Scaling laws for synthetic data}
Conceptually, there is no reason to restrict our approach
to 
a finite dataset of synthetic images. We could devise a training process which sees each image only once
~\cite{parisi2019continual}.

Yet, despite this scaling potential, the quality of the resulting classifier is bounded by the expressivity of the generator and the concepts it can reliably reproduce. No matter how intriguing the promise of an ``infinite dataset'' via data generation might be, practical applications 
are bound by costs linked to computation and storage, as well as the moderation of the content fueling this generator. The latter has strong implications we discuss next.

\mypar{Data and model bias}
Because of its pioneering role as a source of images to train generic models, and all it has done to advance the computer vision field, ImageNet and some of its bias has been under heavy scrutiny~\cite{denton2021genealogy,luccioni2022bugs}. 
Its synthetic counterparts have no reason to be immune to bias.

\looseness=-1
The main advantage of training with synthetic dataset is also its biggest flaw. Instead of manually curating and annotating a dataset, this process is outsourced to a text-to-image generator, whose training data is not always known. Our study is based on the text-to-image generator of Stable Diffusion (SD). SD is trained on LAION-2B~\cite{schuhmann2022laion}, a dataset scraped from the internet and 
filtered in an automatic way using CLIP~\cite{radford2021clip}.
LAION has been shown to contain problematic content~\cite{birhane21multimodal}
and SD models to memorize at least part of the training set~\cite{carlini2023extracting,somepalli2022diffusion}.
Algorithmic bias is not only due to bias in the data~\cite{hooker2021moving}, yet biased datasets lead to biased models and predictions~\cite{steed2021image,aka2021measuring,salman2022when}. 
Frameworks such as~\cite{hutchinson2021towards} could be considered to increase transparency and accountability.

On top of the bias in the data, the architecture itself constraints the generated images, and as such, propagates and potentially amplifies~\cite{bianchi22easily} existing bias.
A major one that we have discussed earlier is the lack of diversity. An obvious corollary is the fact that stereotypes are reinforced.
The options we have explored mitigate this issue to some limited extent, in that it improves classification results, but this issue is far from being solved.
Finally, there are many societal implications of using such models to generate synthetic datasets for training computer vision models, and a more thorough and multi-disciplinary discussion is required.

%% file: tex/7_conclusions.tex
\section{Conclusions}
\label{sec:conclusions}

In this paper, we study to which extent ImageNet, arguably the most popular computer vision dataset, can be replaced by a dataset synthesized by a
text-to-image generator.
Through an extensive study, we find that one can learn models that exhibit
surprisingly good performance on fine-grained classification tasks like \imnethlong and \imnetlong  without any class-specific prompting. However, the most important result of this study is the finding that models trained on synthetic data exhibit exceptional generalization capability
that rivals with models learned with real images. We see this study as merely a first glimpse of what is now possible with the latest large models in terms of visual representation learning.
We envision that similar approaches could be used to fine-tune or adapt models, using those synthetic datasets side-by-side with real ones.

%% file: tex/99_supplementary.tex
\section{Implementation details}\label{sec:implementation_details}

In all experiments the encoder $f_\theta$ is a ResNet50~\cite{he2016resnet} encoder, trained for 100 epochs (unless otherwise stated) with mixed precision in PyTorch~\cite{pytorch} using 4 GPUs where batch norm layers are synchronized.
We use an SGD optimizer with 0.9 momentum, a batch size of 256 and a learning rate linearly increased during the first $10\%$ of the iterations and then decayed with a cosine schedule.
Unless otherwise stated, we use the data augmentation pipeline from DINO~\cite{caron2021dino} with 1 global and 8 local crops ($M_g$~=~1 and $M_l$ = 8).
For Stable Diffusion we use 50 diffusion steps and a guidance scale factor of 7.5 for all experiments. We generate RGB images of size $512 \times 384$.

\section{Evaluation protocol}\label{sec:evaluation_protocols}

We evaluate our models in two ways.
For the different ImageNet test sets, \ie, datasets with images from the training classes (ImageNet-Val/v2/R/A/Sketch), we use the pretrained models as well as the classifiers we learn during pretraining with synthetic images.
For the classification tasks on novel classes, \ie, on the 10 small transfer datasets considered in Tab.~2 of the main paper plus the ImageNet-CoG benchmark in~\Cref{sec:cog},
we freeze the pretrained encoder and train from scratch a new set of linear classifiers for each transfer task.
The list of all datasets we use is given in~\Cref{tab:datasets}.

For transfer learning evaluations, we follow the linear classification protocols from~\cite{kornblith2019transfer,sariyildiz2021cog}.
More precisely, for each of the transfer datasets, we first extract image representations (features) from the pretrained encoders and then train linear logistic regression classifiers using these features.
For the larger transfer datasets, \ie, iNaturalist 2018~\cite{van2018inaturalist} and iNaturalist 2019~\cite{van2018inaturalist} datasets and the CoG levels, we train linear classifiers in PyTorch~\cite{pytorch} using SGD, following~\cite{sariyildiz2021cog}.
For the remaining 8 smaller transfer datasets, we follow~\cite{kornblith2019transfer} and train classifiers using L-BFGS implemented in Scikit-learn~\cite{scikitlearn}.
In all cases, we resize the images with bicubic interpolation so that their shortest side is $224$ pixels, and then take a central crop of $224 \times 224$ pixels.
We tune hyper-parameters (learning rate and weight decay for the SGD optimizer, and regularization coefficient for the L-BFGS optimizer) using Optuna~\cite{optuna2019} over at least 25 trials.
Code for evaluations can be found here\footnote{\resizebox{0.92\linewidth}{!}{\url{https://github.com/naver/trex/tree/master/transfer}}}.

\input{tab/datasets.tex}

\section{Extended experimental results}

\subsection{Impact of data augmentation}
\label{sec:augmentation}

We conducted some basic experiments to evaluate the impact of different data augmentation strategies when learning from synthetic datasets.
In~\Cref{tab:augmentations}, we report the performance of
models trained on the simplest variant of \imnethsdlong, \ie, using the class name as the prompt, utilizing either PyTorch~\cite{torchvision,pytorch} or DINO~\cite{caron2021dino} augmentations.
Although the gains for the real images are relatively small (less than one percent), the gains for \imnethsdlong are over $14\%$.
We believe this shows two things: i) Synthetic images can benefit from the same augmentations as real images, and ii) these transformations are good for domain generalization.
Indeed, strong transformations have been shown to improve domain generalization~\cite{volpi2021cvpr}, and consequently can reduce the  sim-to-real gap.

\input{tab/im100_aug}

\subsection{Results on the ImageNet-CoG~\cite{sariyildiz2021cog} benchmark}
\label{sec:cog}

\input{tab/cog.tex}

We also evaluated our best ImageNet-SD model on the ImageNet-CoG benchmark introduced in~\cite{sariyildiz2021cog} to measure concept generalization.
This benchmark consists of evaluations on the set of training classes of ImageNet-1K (\imnet) and five ``concept generalization levels'', \ie, five \imnet-size datasets of 1000 concepts each.
These 5 concept generalization levels contain concepts from the full ImageNet-19K dataset which do not appear in \imnet.
Moreover, they are ordered, \ie, each containing concepts that are semantically further and further from the \imnet{} ones.

We follow the evaluation protocol presented in~\Cref{sec:evaluation_protocols} and report Top-1 accuracy obtained on the test sets of these datasets in~\Cref{tab:cog}.
We compare the performance of the best \imnetsdlong{} model (from~Tab.~2 of the main paper) to strong baselines trained on \imnetlong{} like the supervised RSB-A1~\cite{wightman2021rsb} and self-supervised DINO~\cite{caron2021dino} models.
We observe that on $L_5$, which is the most challenging level, the performance of the representations learned on synthetic images is comparable to that of learned on real images.
As we move towards $L_1$, we see that the gap between these two models increases in favor of RSB-A1.
Finally, after training classifiers (only) using the real images of \imnet, our model reaches $70.4\%$ accuracy, significantly closing the gap to even the most optimized models trained on real data like RSB-A1.

\subsection{Analysis of the learned features}
\label{sec:feature_analysis}

\input{fig/supp_feature_analyses.tex}

In this section, we analyze and contrast the \textit{representations} obtained with models we trained using synthetic images to representations from models trained on real images.
For this analysis, we used ImageNet-SD models for images that were generated using the default prompt guidance scale of Stable Diffusion, \ie, 7.5.
We perform our analysis for \imnethlong and using \textbf{four metrics}: i) Sparsity, ii) intra-class distance, iii) feature redundancy and iv) coding length.
Note that we use the terms ``representations'' and ``features'' interchangeably.

We compare four different models trained on either real or synthetic data for the 100 classes of \imnethlong: One model trained on real images, ImageNet-100-Real, two models trained on synthetic image sets of the same size obtained by using two different prompts: $p_c = $ ``$c$'' and $p_c = $ ``$c, h_c$ inside $b$'', and the ImageNet-100-SD-10x model, trained using ten times more images.

We perform these analyses on all the datasets listed in~\Cref{tab:datasets}, except for the 5 ImageNet-CoG levels.
For the sake of this study, we split them into three groups:
i) ImageNet-100-Val/v2, ii) ImageNet-100-Sketch/A/R and iii) the 10 transfer datasets (long-tail and small-scale).
For each pretrained model and dataset, we extract features for either only the images in the test set (for the ImageNet test sets), or for all images (for the small transfer datasets).
We then compute each of the four metrics separately on each dataset, and average them over all datasets in the same group.
Before computing metrics, we $\ell_2$-normalize features.

Result analysis for each of the four metrics follows.

\looseness=-1
\mypar{Sparsity}
Inspired by~\cite{kornblith2021why}, we compute feature {\em sparsity ratio}, \ie, the percentage of feature dimensions close to zero with a threshold of $10^{-5}$.
We report sparsity ratios in~\Cref{fig:sparsity}.
We see that the sparsity ratio for the models trained on synthetic images increases as the ``diversity'' of a synthetic dataset increases, \ie, we see gradual increase in sparsity scores from $p_c = $ ``$c$'' and $p_c = $ ``$c, h_c$ inside $b$'' to ImageNet-100-SD-10x.
This observation aligns with their performance as well, \ie, in the main paper we show that ImageNet-100-SD-10x performs best in general while $p_c = $ ``$c$'' performs worst.
More interestingly, we see that ImageNet-100-Real, the model trained on real images, learns the most sparse representations.

\looseness=-1
\mypar{Intra-class distance}
In the main paper, we
present
simple ways to increase the diversity of synthetic images.
Now we check if these efforts increase the variance of samples in the representation space.
To do that, we compute the average $\ell_2$-distance between samples from the same class (\ie, intra-class distance).
We see in~\Cref{fig:intraclass} that models trained with more diverse images indeed learn representations with higher intra-class variance.

\looseness=-1
\mypar{Feature redundancy}
Following~\cite{wang2022revisiting}, we compute feature redundancy, \ie, average pairwise Pearson correlation among dimensions.
From~\Cref{fig:feature_redundancy} we see that the redundancy of features learned on real images increase more rapidly than the ones learned on synthetic images, as we move from ImageNet-100-Val/v2 towards out-of-domain or transfer datasets.

\looseness=-1
\mypar{Coding length}
To further investigate our observation on feature redundancy, we follow~\cite{yu2020learning} and compute the average coding length per sample on each dataset (see \Cref{fig:coding_length}).
We see that models trained on ImageNet-100-Real and ImageNet-100-SD-10x are comparable.

\subsection{Impact of guidance scale and diffusion steps}
\label{sec:scale_steps_photo_of_ablation}

\looseness=-1
In \Cref{fig:scale_steps_ablation} we analyse the impact of the guidance scale and diffusion step hyper-parameters of Stable Diffusion~\cite{rombach2022high}.
As we discuss in the main paper, a lower guidance scale leads to more visual diversity and that is reflected of performance.
Values of 1 to 3 all seem like a good choice.
When it comes to the number of diffusion steps, values like 25 and (the default) 50 seem like a safe choice, with 25 being slightly worse, but requiring half the time to extract.
Interestingly, using more steps seems to slightly hurt performance on the training classes.
It is worth noting that transfer learning performance is surprisingly and consistently high for even 5 diffusion steps.
This corroborates recent finding that training on complex but possibly semantically meaningless images like fractals~\cite{KataokaIJCV2022} or sinusoidal waves~\cite{takashima2023visual} can provide a strong starting point for visual representations that generalize well.

\begin{figure*}[t]
    \centering
    \begin{subfigure}[t]{.48\linewidth}
        \centering
        \input{fig/supplementary/guidance_scale_ablation}
        \caption{Impact of the guidance scale parameter}
        \label{fig:scale_ablation}
    \end{subfigure}
    ~
    \begin{subfigure}[t]{.48\linewidth}
        \centering
        \input{fig/supplementary/steps_ablation}
        \caption{Impact of the nubmer of diffusion steps}
        \label{fig:steps_ablation}
    \end{subfigure}
    \caption{{\bf Impact of the guidance scale parameter and number of diffusion steps.} Top-1 Accuracy on ImageNet-100 and averaged over 10 transfer datasets for $p_c = $~``$c$, $d_c$''. In the left plot, steps are set to 50, in the right plot guidance scale is 7.5.}
    \label{fig:scale_steps_ablation}
\end{figure*}

\subsection{Prefixing the prompt with domain identifiers}

Handcrafted, dataset-level prompt engineering was used for the zero-shot experiments in the CLIP~\cite{radford2021clip} paper.
For example they use the prompt template ``A photo of a $c$'' as default for classification tasks. For other fine-grained image classification datasets they go one step further and append ``a type of \{domain\}'' where \{domain\}=\{pet,food,aircraft\} for datasets containing pet, food or aircraft classes.

In the main paper, instead presented automatic ways of clarifying the domain, \ie, using extra information from WordNet for each class.
In~\Cref{tab:photo_of_image_of} we present some preliminary results when using generic prompt templates like ``a photo of $c$'' and ``an image of $c$'' as input to the Stable Diffusion v1.4 model.
We found them to decrease performance for \imnethlong.

\begin{table}[h]
    \centering
    \resizebox{\linewidth}{!}{
    \begin{tabular}{lccc}
    \toprule
    $p_c$~ & ``$c$'' & ``a photo of $c$'' & ``an image of $c$'' \\ \midrule
    Top-1 Acc. & \textbf{64.8} & 59.5 & 58.3 \\ \bottomrule
    \end{tabular}
    }
    \caption{Top-1 Accuracy on ImageNet-100 when prepending the prompt with domain identifiers. Guidance scale is equal to 2.0.}
    \label{tab:photo_of_image_of}
\end{table}

\subsection{Additional scaling plots for synthetic data}

\begin{figure*}
    \centering
    \begin{subfigure}[t]{.48\linewidth}
        \centering
        \input{fig/scaling_plot}
        \caption{Top-1 accuracy on \imnethlong.}
        \label{fig:supp_scaling_imnet}
    \end{subfigure}
    ~
    \begin{subfigure}[t]{.48\linewidth}
        \centering
        \input{fig/scaling_plot_transfer_supp}
        \caption{Top-1 accuracy on the 10 transfer datasets from~\Cref{tab:transfer} of the main paper.}
        \label{fig:supp_scaling_transfer}
    \end{subfigure}
    \caption{{\bf Scaling the number of training images.}
            Accuracy when training on \imnethlong using $(1/10)$-th to 50$\times$ images (relative to the real dataset size).
            \Cref{fig:supp_scaling_transfer} is also shown in the main paper.}
    \label{fig:scaling_supp}
\end{figure*}

In \Cref{fig:scaling_supp} we report accuracy when training on \imnethlong using $(1/10)$-th to 50$\times$ images, relative to the real dataset size.
~\Cref{fig:supp_scaling_imnet} suggests that generating more images with basic prompts might not be enough, and that a performance leap will require advanced prompt engineering.
We consider a study on scaling synthetic datasets is important, but beyond the scope of this paper.
Note that~\Cref{fig:supp_scaling_transfer} is also shown in the main paper and repeated here for completeness.

\subsection{Additional spider plots}

In~\Cref{fig:four_spiders} we show spider plots for the models trained on either real or synthetic data for \imnethlong and \imnetlong.
In both cases,
we show two plots which respectively report
top-1 and top-5 accuracy for the ImageNet datasets, \ie, ImageNet-Val/v2/R/A/Sketch.
For transfer datasets and similar to the teaser figure in the main paper, we report top-1 accuracy averaged over the transfer datasets in each of the following three groups: \textbf{(a)} eight common small-scale datasets (Aircraft~\cite{maji2013aircraft},
Cars196~\cite{krause2013cars},
DTD~\cite{cimpoi2014texture},
EuroSAT~\cite{helber2019eurosat},
Flowers~\cite{nilsback2008flowers},
Pets~\cite{parkhi2012pets},
Food101~\cite{bossard2014food101},
SUN397~\cite{xiao2010sun}), \textbf{(b)} two long-tail datasets (iNat2018~\cite{van2018inaturalist} and iNat2019~\cite{van2018inaturalist}), and \textbf{(c)} the five datasets (``levels'') of the CoG benchmark~\cite{sariyildiz2021cog}.

\input{fig/supp_spider.tex}

\section{Extended qualitative results}
\label{sec:sup_qualitative}

\looseness=-1

In this section, we provide additional qualitative results.
First we show random images for {\em all} \imnethlong classes from three datasets: \imnethlong-Val (real images) and two \imnethsdlong datasets generated by the prompts $p_c=$ ``$c$'' and $p_c=$ ``$c$, $h_c$ inside $b$''.
Then we discuss in more detail several types of issues that we observed in these synthetic images. Unless otherwise stated, the guidance scale used is 7.5.

\mypar{Qualitative results for \textit{all} ImageNet-100 classes}
In~\Cref{fig:images_for_im100_classes}, we show a few random images from each of the 100 classes in ImageNet-100, for three datasets: i) The real images from ImageNet-100, ii) synthetic images generated by a simple prompt, which is only composed of the name of the class,
and iii) synthetic images generated with {guidance scale equal to 2.0 and} a prompt that enforces thoses classes to appear in diverse backgrounds
to improve the diversity of generated images. From this exhaustive list, even with a few images per class, one can observe a number of issues around the semantics, diversity and domain of those images.

\mypar{Showcasing domain and diversity issues}
We also show extended results for three classes in order to illustrate
issues related to the domain and diversity.
\Cref{fig:crabs} compares generated images between two fine-grained classes of crabs, while~\cref{fig:shih_tsu} shows many images from multiple different generated datasets for a single dog class.
We discuss both figures in the next subsections.

\subsection{Semantic errors}

From closely inspecting the generated images
we can see that there exists two classes for which the prompt $p_c = $~``$c$'' produces images of the wrong semantics: For the classes ``papillon'' and ``wing'', we see the generated images in the middle column of~\Cref{fig:images_for_im100_classes} to be wrong due to \textit{polysemy} associated with the class names.
What is more, although not fully visible from the small set of images we show here, we saw that semantics are partially wrong for at least the classes ``green mamba'', ``walking stick'' and ``iron''.
For ``green mamba'', although the synset refers to the snake species, there is a car model of the same name appearing in some of the generated images instead.
For ``walking stick'', the synset refers to the insect, while a subset of the generated images also contained walking sticks that are not insects.

As we discuss in the paper, appending the hypernym or definition of each synset seems to fix polysemy issues in many cases, including the ones mentioned above.
However, we can see at least two cases where adding the hypernym in the prompt leads to worse results.
According to WordNet~\cite{miller1995wordnet}, the hypernym for ``shih-tzu'' is ``toy dog'' something that results in dog-shaped toys in many of the generated images (see also~\Cref{fig:shih_tsu}).
Another example is the class ``boathouse'', where appending the parent class ``shed'' leads to sheds that are not inside a body of water.

\subsection{NSFW content}
\label{sec:nsfw}

Another issue that was not very prominent, but still visible, even in the case of generic animal and object categories present in ImageNet-100, was the fact that some of the generated images contained NSFW (Not Suitable For Work) content in the form of nudity.
The open-source code for Stable Diffusion comes with a highly selective safety module, that discards generated images that might contain NSFW content.\footnote{{\url{https://huggingface.co/CompVis/stable-diffusion-v1-4?text=Safety}}}
We disabled this module when generating images for the ImageNet synsets as we wanted to study the model as-is first, and to understand the problem.

We thoroughly inspected all classes of ImageNet-100 and observed minor NSFW issues with two of the classes: 1)
The basic prompt for the class ``sarong'' led to a few images that had partial nudity.
This effect was exaggerated when adding the description of the concept that reads ``a loose skirt consisting of brightly colored fabric wrapped around the body; worn by both women and men in the South Pacific".
It seems that words like ``body'' biases the image generation process towards more NSFW content.
2) Prompts for the class ``ski mask'' in combination with certain backgrounds from the Places dataset~\cite{zhou2017places} also resulted in nudity.
Overall, we want to emphasize that the Stable Diffusion models we tested were all highly susceptible to generate such content.

\subsection{Misrepresentation of biodiversity}

The degree of misrepresentation of biodiversity in the images generated from Stable Diffusion is very high.
We partially showcase the issue in~\Cref{fig:crabs} where we show many generated images for two fine-grained classes, \ie, ``rock crab'' and ``fiddler crab''.

``Rock crab'' is defined in WordNet as ``crab of eastern coast of North America'', while the ``fiddler crab'' as a ``burrowing crab of American coastal regions having one claw much enlarged in the male''.
The fact that the male fiddler crab has one claw much larger is a prominent theme when it comes to the real ImageNet-100 images shown on the right side of~\cref{fig:crabs_real}.

It does not take an expert ecologist to see that, although most of the generated images capture the coarser class ``crab'', the visual differences between the two sets of images, \eg, in~\cref{fig:crabs_base}, are not focusing on the single enlarged claw for the fiddler crab case.
What is more, the exhibited intra-class visual diversity, \ie, crabs of different shapes and colors, seems to exceed a single species of crab.

This is just a single example, but from our inspection of many other fine-grained animal and fungi classes, we could see that this is not an isolated issue. On the contrary, it seems prominent across many fine-grained domains.
One exception for the subset of ImageNet classes we delved into is dog breeds, possibly due to the sheer volume of dog images on the internet.
It is however fair to say that the generated images highly misrepresent biodiversity.

It is worth noting that, as Luccioni and Rolnick discuss in their recent paper~\cite{luccioni2022bugs}, the ImageNet dataset itself contains a number of issues when it comes to the annotations of fine-grained classes of wild animals.
They found that ``many of the classes are ill-defined or overlapping, and that $12\%$ of the images are incorrectly labeled, with some classes having $> 90\%$ of images incorrect''.
Although we did not conduct a similar experiment using experts, we expect similar statistics to be much higher for the images generated by Stable Diffusion.

\subsection{Semantic issues arising with backgrounds}

A common issue we observe when adding diverse backgrounds to class images is that a subset of the generated images do not really contain the object, and merely reflect the background scene.
See for example the images in the first and last row, on the last column of~\Cref{fig:crabs_bg}, and a few more spread in that figure, or the background samples for class ``reel'' in~\Cref{fig:images_for_im100_classes}.
This is to be expected given how a prompt like this is relying on the compositionality of the Stable Diffusion model.

What is really interesting is that in some cases the resulting images, although not containing an instance from the class, retains some of the object's shape or texture in the background.
See for example a pedestal-looking table in~\Cref{fig:crabs_bg} for class ``pedestal'', a pirate themed bedroom for class ``pirate'', green shirts for ``green mamba'', or the red-ish produce stand for ``red fox''.

\subsection{Issues with diversity}

We observe issues with diversity for most of the classes when only the class name is used as the prompt, \eg, in the middle set of results in~\cref{fig:images_for_im100_classes}.
This is also visible for the crab classes in~\cref{fig:crabs_base}, or the Shih-tzu class in~\Cref{fig:shih_tsu_base}, \cref{fig:shih_tsu_d} and \cref{fig:shih_tsu_bg}.
We see that such issues are partially solved when {lowering the guidance scale and relying less to the prompt, or} using  %
backgrounds (\eg, the right-most set of images in~\cref{fig:images_for_im100_classes}).
We expect more advanced prompt engineering to further increase diversity.

As expected, increasing diversity correlates with more semantic errors.
We see that such issues appear far more frequently in the most diverse synthetic dataset, \ie, as shown in the right-most set of images of~\cref{fig:images_for_im100_classes}.

\subsection{Non-natural images}

Even from the very small random sample of generated images shown in the figures of this paper,
we see that there is a non-negligible percentage of the generated images that are non-natural.
They can be illustrations, graphics images or even paintings.
This is not necessarily undesirable and it can lead to models with higher robustness to related domain changes.

\subsection{Varying the stable diffusion parameters}
\label{sec:sd_params}

We identify two important parameters for Stable Diffusion, which affect the visual quality of generated images: The guidance scale and the number of diffusion steps.
In~\Cref{fig:sd_params} we show several examples where we vary one of these two parameters.
More specifically, we generate images for the ImageNet synset \texttt{n01558993} with class name ``robin, American robin, Turdus migratorius'', for the simplest case where the prompt is just the class name.
We fix the seed to 1947262 and vary either the guidance scale or the number of diffusion steps.

\mypar{Guidance Scale}
From~\Cref{fig:sd_scale}, we see that increasing the guidance scale coefficient over 10 starts giving hyper-realistic results.
When the scale is under 2, we see that many details of the class are not really prominent.

\mypar{Diffusion Steps}
From~\Cref{fig:sd_steps}, we see that, although with 5 steps the generated images still contain a lot of noise, running 25-50 steps is enough for fully-formed, sharp images to emerge.
Since this is a parameter that linearly impacts generation time, increasing the number of steps further than 50 seems excessive.

\mypar{Output Resolution}
The resolution that was used during training of the Stable Diffusion models was ($512 \times 512$).\footnote{\url{https://github.com/CompVis/stable-diffusion}}
We notice that if one deviates from this training resolution, generated results get worse.
We chose to simply switch the aspect ratio to the one for the average ImageNet image and keep the long dimension to 512.

\input{fig/sd_params}

\input{fig/shih-tzu}

\input{fig/crabs}

\input{fig/supp_qualitative_per_class.tex}

%% file: tab/datasets.tex
{
\begin{table*}[t]
    \centering
    \small
    \adjustbox{max width=\textwidth}{
    \begin{tabular}{@{}lrrrrcc@{}}
    \toprule
    \multicolumn{1}{l}{Dataset} &
        \multicolumn{1}{c}{\# Classes} &
        \multicolumn{1}{c}{\begin{tabular}[c]{@{}c@{}}\# Train\\ samples\end{tabular}} &
        \multicolumn{1}{c}{\begin{tabular}[c]{@{}c@{}}\# Val\\ samples\end{tabular}} &
        \multicolumn{1}{c}{\begin{tabular}[c]{@{}c@{}}\# Test\\ samples\end{tabular}} &
        \multicolumn{1}{c}{\begin{tabular}[c]{@{}c@{}}Val\\ provided\end{tabular}} &
        \multicolumn{1}{c}{\begin{tabular}[c]{@{}c@{}}Test\\ provided\end{tabular}} \\
    \toprule
    \multicolumn{7}{c}{{\em ImageNet test sets (training classes)}} \\
    ImageNet-Val~\cite{russakovsky2015ilsvrc} (IN-Val)    & 1000  & \multicolumn{1}{c}{--} & \multicolumn{1}{c}{--} & 50000            & --  & $\checkmark$ \\
    ImageNet-v2~\cite{recht2019imagenet} (IN-v2)          & 1000  & \multicolumn{1}{c}{--} & \multicolumn{1}{c}{--} & $3 \times 10000$ & --  & $\checkmark$ \\
    ImageNet-Sketch~\cite{wang2019learning} (IN-Sketch)   & 1000  & \multicolumn{1}{c}{--} & \multicolumn{1}{c}{--} & 50889            & --  & $\checkmark$ \\
    ImageNet-R~\cite{hendrycks2021many} (IN-R)            & 200   & \multicolumn{1}{c}{--} & \multicolumn{1}{c}{--} & 30000            & --  & $\checkmark$ \\
    ImageNet-A~\cite{hendrycks2021natural} (IN-A)         & 200   & \multicolumn{1}{c}{--} & \multicolumn{1}{c}{--} & 7500             & --  & $\checkmark$ \\
    \midrule
    \multicolumn{7}{c}{{\em Transfer tasks (novel classes)}} \\
    Aircraft~\cite{maji2013aircraft}            & 100   & 3334   & 3333  & 3333   & $\checkmark$ & $\checkmark$ \\
    Cars196~\cite{krause2013cars}               & 196   & 5700   & 2444  & 8041   & --           & $\checkmark$ \\
    DTD~\cite{cimpoi2014texture}                & 47    & 1880   & 1880  & 1880   & $\checkmark$ & $\checkmark$ \\
    EuroSAT~\cite{helber2019eurosat}            & 10    & 13500  & 5400  & 8100   & --           & --           \\
    Flowers~\cite{nilsback2008flowers}          & 102   & 1020   & 1020  & 6149   & $\checkmark$ & $\checkmark$ \\
    Pets~\cite{parkhi2012pets}                  & 37    & 2570   & 1110  & 3669   & --           & $\checkmark$ \\
    Food101~\cite{bossard2014food101}           & 101   & 68175  & 7575  & 25250  & --           & $\checkmark$ \\
    Pets~\cite{parkhi2012pets}                  & 397   & 15880  & 3970  & 19850  & --           & $\checkmark$ \\
    iNaturalist 2018~\cite{van2018inaturalist}  & 8142  & 437513 & -- & 24426 & -- & $\checkmark$ \\
    iNaturalist 2019~\cite{van2018inaturalist}  & 1010  & 265213 & -- & 3030  & -- & $\checkmark$ \\
    CoG $L_1$ \cite{sariyildiz2021cog}          & 1000  & 895359 & 223445 & 50000 & --           & $\checkmark$ \\
    CoG $L_2$ \cite{sariyildiz2021cog}          & 1000  & 892974 & 222814 & 50000 & --           & $\checkmark$ \\
    CoG $L_3$ \cite{sariyildiz2021cog}          & 1000  & 876495 & 218708 & 50000 & --           & $\checkmark$ \\
    CoG $L_4$ \cite{sariyildiz2021cog}          & 1000  & 886013 & 221115 & 50000 & --           & $\checkmark$ \\
    CoG $L_5$ \cite{sariyildiz2021cog}          & 1000  & 873630 & 218024 & 50000 & --           & $\checkmark$ \\
    \bottomrule
    \end{tabular}
    }
    \caption{
        {\bf Datasets} we use for evaluating our models.
    }
    \label{tab:datasets}
\end{table*}
}

%% file: tab/im100_aug.tex
\begin{table}[!h]
    \centering
    \resizebox{\linewidth}{!}{
        \begin{tabular}{lrr}
            \toprule
             Training Dataset  & PyTorch~\cite{pytorch} &  DINO (+ Multi-crop) \\
            \toprule
             \textit{\imnethlong (real)}  & \textit{86.6} & \textit{87.4} \diffup{0.80} \\
             \imnethsdlong (synthetic) & 28.4 & 43.1 \textbf{\diffup{14.6}}\\
             \bottomrule
        \end{tabular}
    }
    \caption{
        {\bf Impact of data-augmentation}
        for models trained on real and synthetic datasets. Performance is measured on the validation set of \imnethlong, \ie on real images.
    }
    \label{tab:augmentations}
\end{table}

%% file: tab/cog.tex
{
\setlength{\tabcolsep}{2pt}
\begin{table}[t]
    \adjustbox{max width=\linewidth}{
    \begin{tabular}{@{}llcccccc@{}}
    \toprule
    Training Dataset & Prompt ($p_c$) / Model & \imnet & $L_1$ & $L_2$ & $L_3$ & $L_4$ & $L_5$ \\
    \toprule
    \multirow{3}{*}{\em \imnetlong}
      & {\em PyTorch~\cite{torchvision}}              & \underline{\em 75.8} & {\em 67.8} & {\em 63.1} & {\em 58.9} & {\em 58.2} & {\em 52.0} \\
      & {\em {RSB-A1}~\cite{wightman2021rsb}}         & {\bf \em 79.8} & \underline{\em 69.9} & \underline{\em 65.0} & \underline{\em 60.9} & \underline{\em 59.3} & \underline{\em 52.8} \\
      & {\em {DINO}~\cite{caron2021dino}}             & {\em 74.8} & {\bf \em 71.1} & {\bf \em 67.2} & {\bf \em 63.2} & {\bf \em 62.6} & {\bf \em 57.6} \\
    \midrule
    \imnetsdlong & $p_c=$ ``$c$, $d_c$''              & 70.4 & 65.7 & 61.8 & 58.5 & 58.0 & {52.4} \\
    \bottomrule
    \end{tabular}
    }
    \caption{
        {\bf Top-1 accuracy on the ImageNet-CoG benchmark~\cite{sariyildiz2021cog}}
        We report performance for the best \imnetsdlong model from~Tab~2. of the main paper (with guidance scale equal to 2).
    }
    \label{tab:cog}
\end{table}
}

%% file: fig/supp_feature_analyses.tex
\begin{figure*}[t]
    \centering
    \begin{subfigure}[t]{.46\linewidth}
        \centering
        \includegraphics[width=\columnwidth]{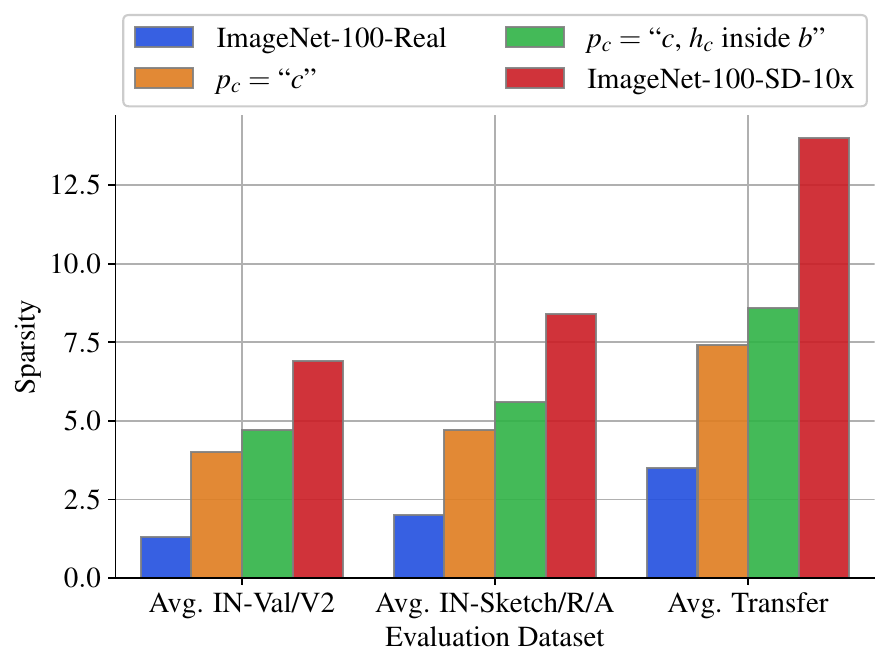}
        \caption{\textbf{Feature Sparsity}}
        \label{fig:sparsity}
    \end{subfigure}
    ~
    \begin{subfigure}[t]{.46\linewidth}
        \centering
        \includegraphics[width=\columnwidth]{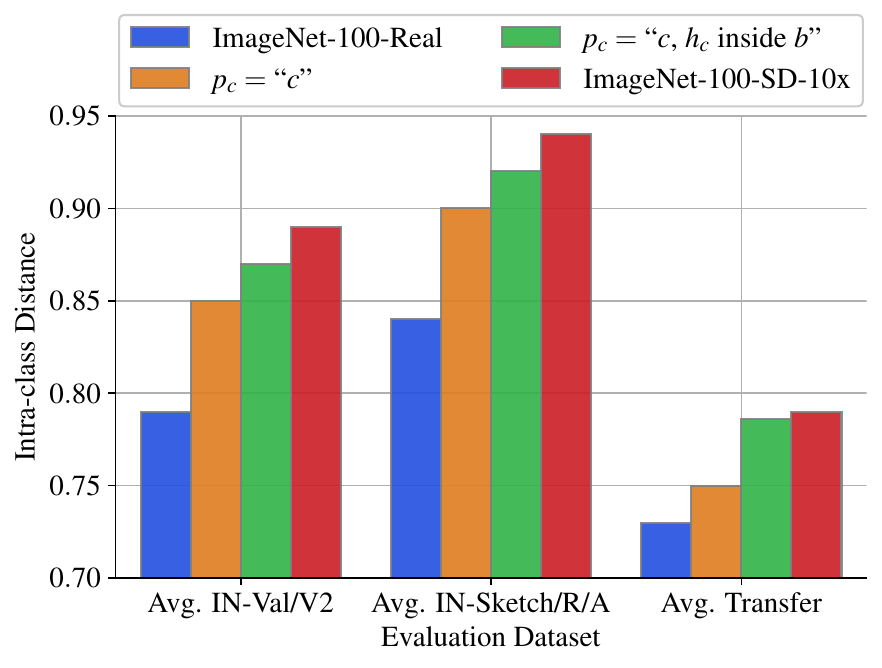}
        \caption{\textbf{Intra-class $\ell_2$-distance}}
        \label{fig:intraclass}
    \end{subfigure}
    \begin{subfigure}[t]{.46\linewidth}
        \centering
        \includegraphics[width=\columnwidth]{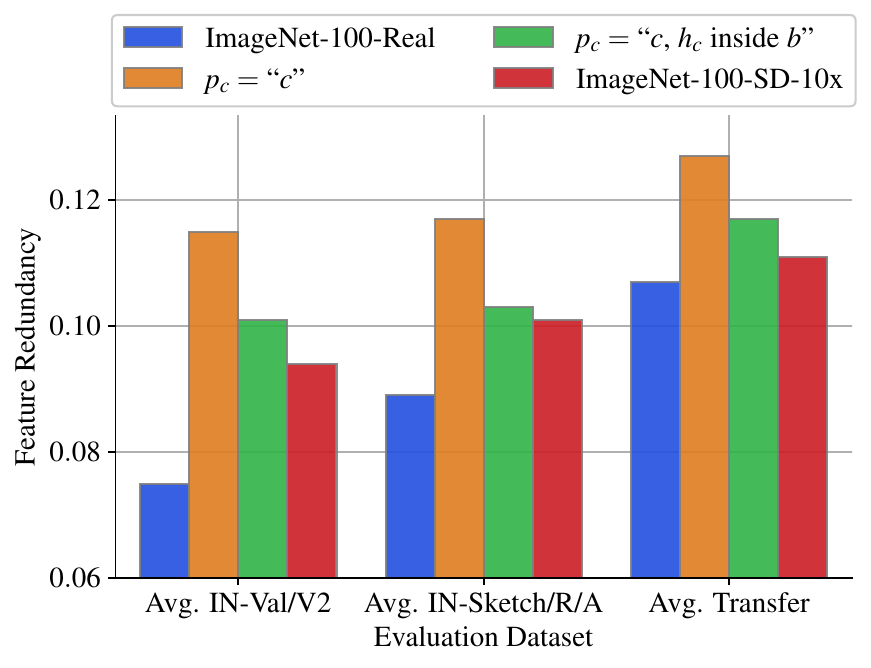}
        \caption{\textbf{Feature redundancy}}
        \label{fig:feature_redundancy}
    \end{subfigure}
    ~
    \begin{subfigure}[t]{.46\linewidth}
        \centering
        \includegraphics[width=\columnwidth]{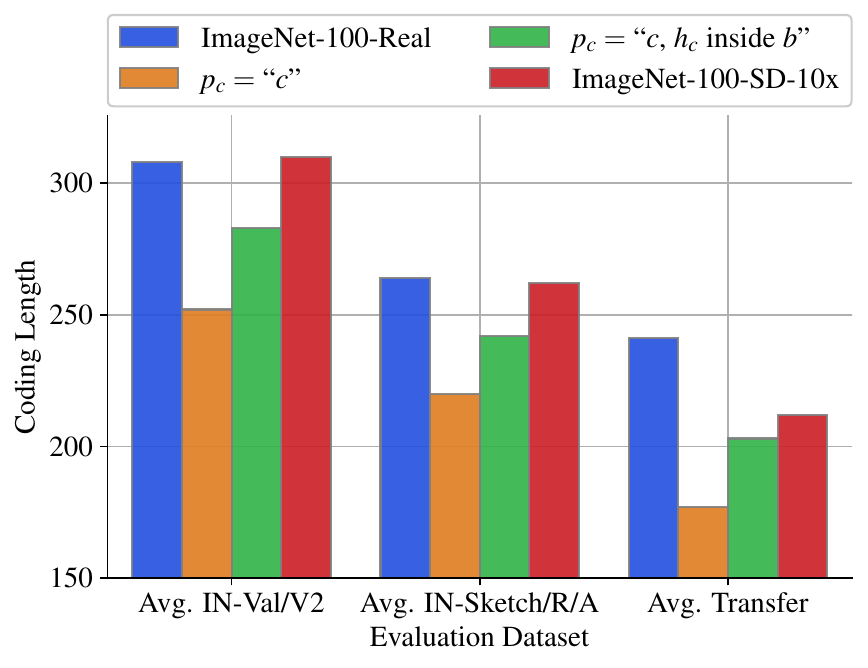}
        \caption{\textbf{Coding length}}
        \label{fig:coding_length}
    \end{subfigure}
    \caption{
        \textbf{Feature analyses} for models.
        We perform these analyses on top of features extracted from pretrained encoders $f$ trained on either real or synthetic data for \imnethlong (training data is specified in the legends of the subfigures).
        For the purpose of this study, we use synthetic data generated with guidance scale equal to 7.5.
        {\em Sparsity} is measured by the percentage of dimensions close to zero~\cite{kornblith2021why}.
        {\em Intra-class $\ell_2$-distance} is the average pairwise $\ell_2$-distance between samples from the same class.
        These two metrics are computed on $\ell_2$-normalized features.
        {\em Feature redundancy}~\cite{wang2022revisiting} is obtained by $\mathcal{R} = \frac{1}{d^2} \sum_i \sum_j | \rho(\mathbf{X}_{:,i}, \mathbf{X}_{:,j}) |$, where $\mathbf{X} \in \real{N \times d}$ is a feature matrix containing $N$ samples, each encoded into a $d$-dimensional representation (2048 in our case) and $\rho(\mathbf{X}_{:,i}, \mathbf{X}_{:,j})$ is the Pearson correlation between a pair of feature dimensions $i$ and $j$.
        {\em Coding length}~\cite{yu2020learning} is measured by $R(\mathbf{X}, \epsilon) = \frac{1}{2} \log \det  (\mathbf{I}_d + \frac{d}{N \epsilon^2} \mathbf{X}^\top \mathbf{X}),$  where $\mathbf{I}_d$ is a $d$-by-$d$ identity matrix, $\epsilon^2$ is the precision parameter set to $0.5$.
    }
    \label{fig:feature_analyses}
\end{figure*}

%% file: fig/supplementary/guidance_scale_ablation.tex
 \begin{tikzpicture}
    \begin{axis}[
        height=5cm,
        xlabel={Guidance Scale},
        ylabel={Accuracy},
        label style={font=\footnotesize},
        xmin=0,
        xmax=10,
        minor y tick num=3,
        xticklabels={0,0.5,1,2,3,7.5,10},
        xtick={1,2,3,4,5,7,9},
        legend pos=south east,
        legend style={font=\small},
        ]

        \addplot[sd1]
            coordinates {
             (1, 0.8)
             (2, 37.8)
             (3, 62.8)
             (4, 64.8)
             (5, 62.2)
             (7, 47.9)
             (9, 43.0)
            }; \leg{ImageNet-100};

        \addplot[sd1t]
            coordinates {
             (1, 7.44)
             (2, 45.1)
             (3, 56.68)
             (4, 56.26)
             (5, 55.0)
             (7, 51.13)
             (9, 50.48)

            }; \leg{Transfer Datasets};

    \end{axis}
\end{tikzpicture}

%% file: fig/supplementary/steps_ablation.tex
 \begin{tikzpicture}
    \begin{axis}[
        height=5cm,
        xlabel={Number of Diffusion Steps},
        ylabel={Accuracy},
        label style={font=\footnotesize},
        xmin=0,
        xmax=110,
        minor y tick num=3,
        xticklabels={5,25,50,100},
        xtick={5,25,50,100},
        legend pos=south east,
        legend style={font=\small},
        ]

        \addplot[sd1]
            coordinates {
             (5, 34.2)
             (25, 47.5)
             (50, 47.9)
             (100, 46.9)
            }; \leg{ImageNet-100};

        \addplot[sd1t]
            coordinates {
            (5, 51.46)
            (25,51.48)
            (50, 51.13)
            (100,51.17)

            }; \leg{Transfer Datasets};

    \end{axis}
\end{tikzpicture}

%% file: fig/scaling_plot.tex
 \begin{tikzpicture}
    \begin{axis}[
        height=5cm,
        xlabel={Number of Images},
        ylabel={Top-1 Accuracy},
        label style={font=\footnotesize},
        xmin=0,
        xmax=10,
        minor y tick num=1,
        xtick={1,3,5,7,8,9},
        xticklabels={0.01$\times$,0.1$\times$,\textbf{1$\times$},10$\times$,20$\times$,50$\times$},
        tick label style={font=\footnotesize},
        legend pos=south east,
        legend style={font=\small},
        ]

        \addplot[real1]
        coordinates {
         (1, 7.3)
         (3, 68.7)
         (5, 87.4)
        }; \leg{real images};

     \addplot[real1, dashed, mark size=0pt]
        coordinates {
         (5, 87.4)
         (7, 87.4)
         (8, 87.4)
         (9,  87.4)
        }; \leg{real images};

    \addplot[sd1]
        coordinates {
         (1, 5.3)
         (3, 47.7)
         (5, 64.8)
         (7, 72.4)
         (8, 72.4)
         (9, 73.3)
        }; \leg{synthetic images};

    \end{axis}
\end{tikzpicture}

%% file: fig/scaling_plot_transfer_supp.tex
 \begin{tikzpicture}
    \begin{axis}[
        height=5cm,
        xlabel={Number of Images},
        ylabel={Top-1 Accuracy},
        label style={font=\footnotesize},
        xmin=2,
        xmax=10,
        minor y tick num=1,
        xtick={3,5,7,8,9},
        xticklabels={0.1$\times$,\textbf{1$\times$},10$\times$,20$\times$,50$\times$},
        tick label style={font=\footnotesize},
        legend pos=south east,
        legend style={font=\small},
        ]

        \addplot[real1]
        coordinates {
         (3, 51.17)
         (5, 58.4)
        }; \leg{real images};

         \addplot[sd1]
        coordinates {
         (3, 51.23)
         (5, 56.26)
         (7, 62.1)
         (8, 62.7)
         (9, 63.2)
        }; \leg{synthetic images};

        \addplot[real5, mark size=0pt]
        coordinates {
         (5, 58.4)
         (7, 58.4)
         (8, 58.4)
         (9,  58.4)
         (10,  58.4)
        };

    \end{axis}
\end{tikzpicture}

%% file: fig/supp_spider.tex
\begin{figure*}
    \centering
    \begin{subfigure}{0.46\linewidth}
        \centering
        \includegraphics[width=\linewidth]{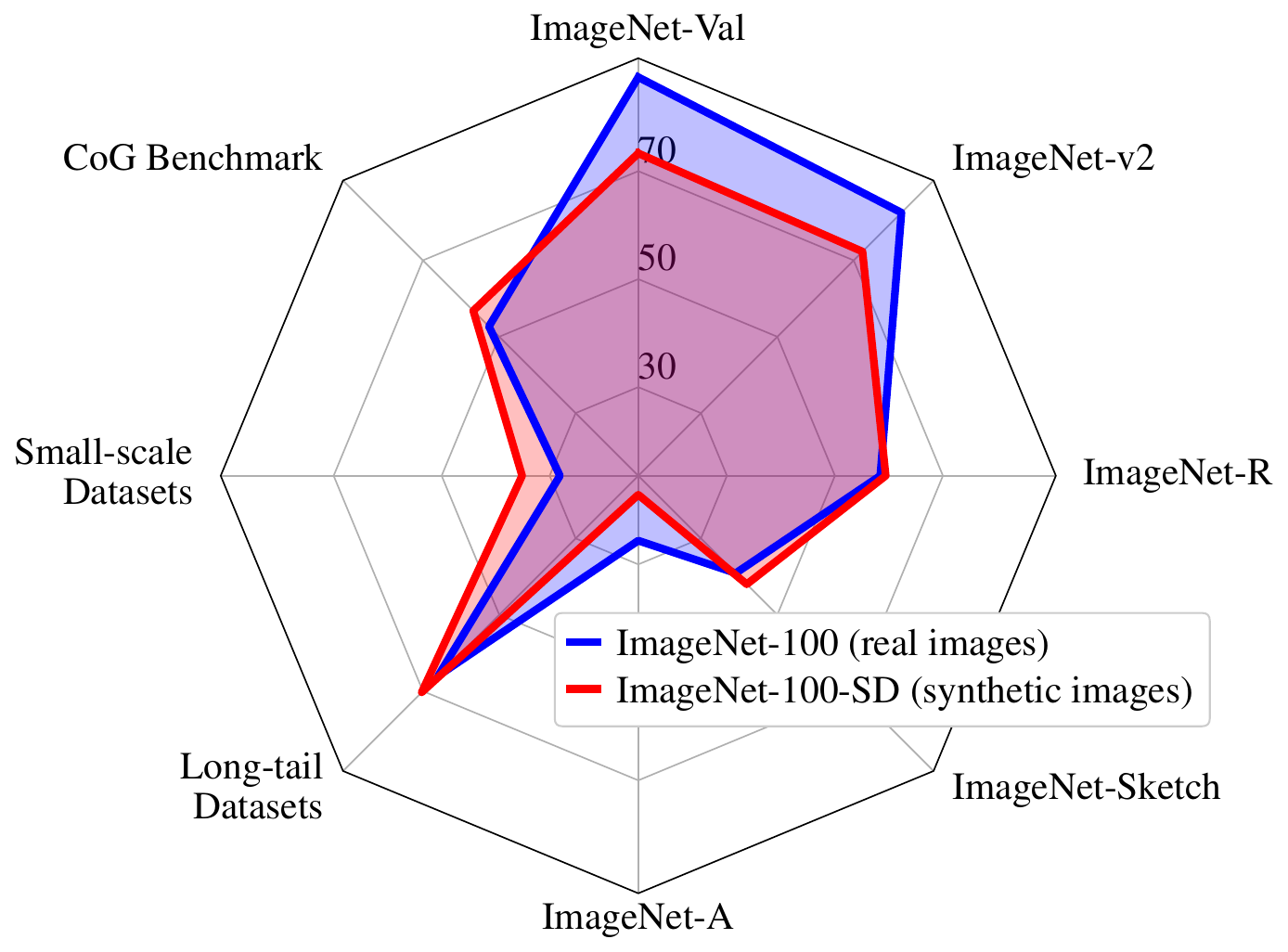}
        \caption{Top-1 accuracy, training on {\bf \imnethlong}.}
        \label{fig:spider_in100_top1}
    \end{subfigure}
    ~
    \begin{subfigure}{0.46\linewidth}
        \centering
        \includegraphics[width=\linewidth]{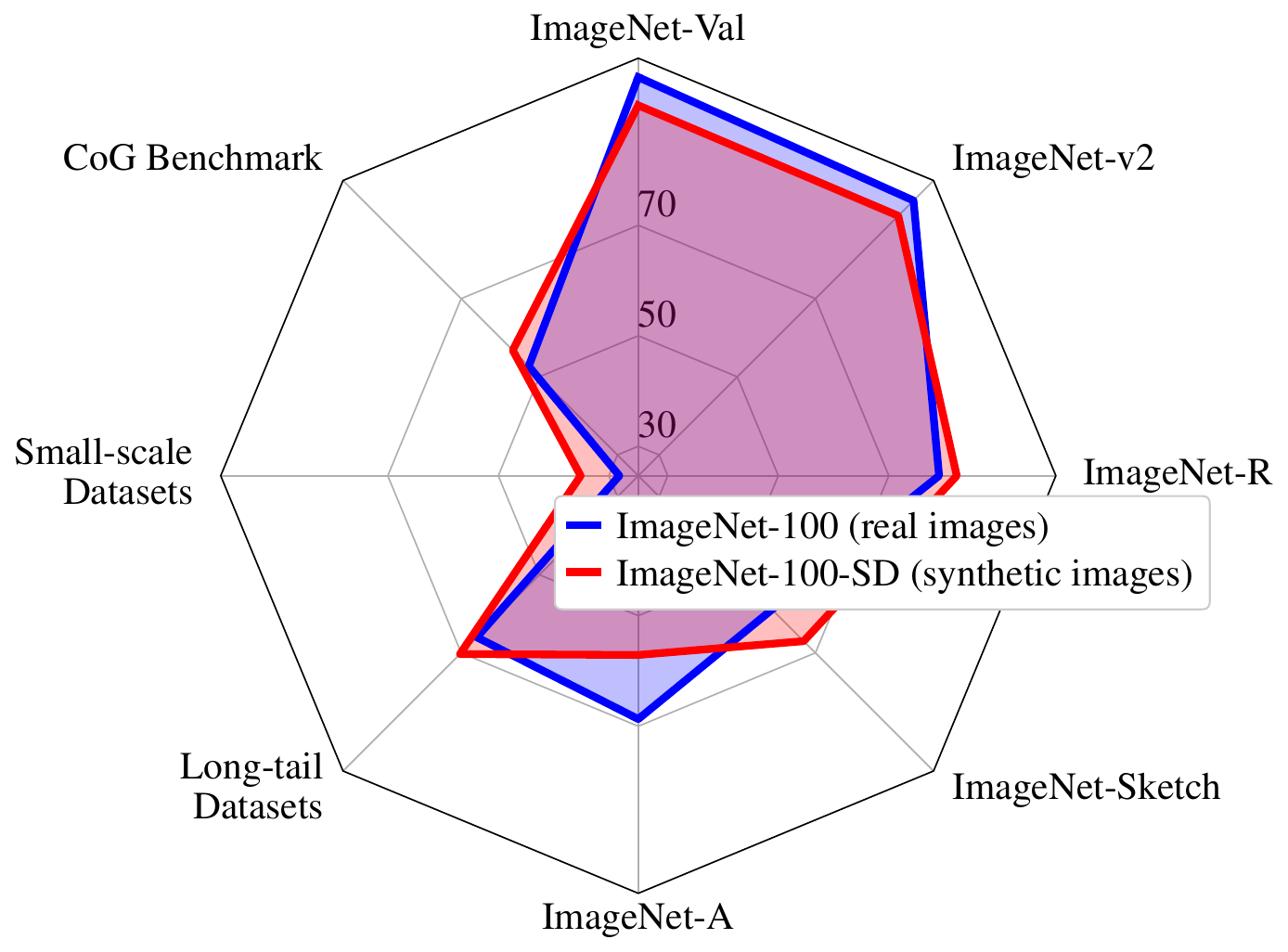}
        \caption{Top-5 accuracy, training on {\bf \imnethlong} (top-1 for transfer tasks).}
        \label{fig:spider_in100_top5}
    \end{subfigure}

    \vspace{15pt}
    \begin{subfigure}{0.46\linewidth}
        \centering
        \includegraphics[width=\linewidth]{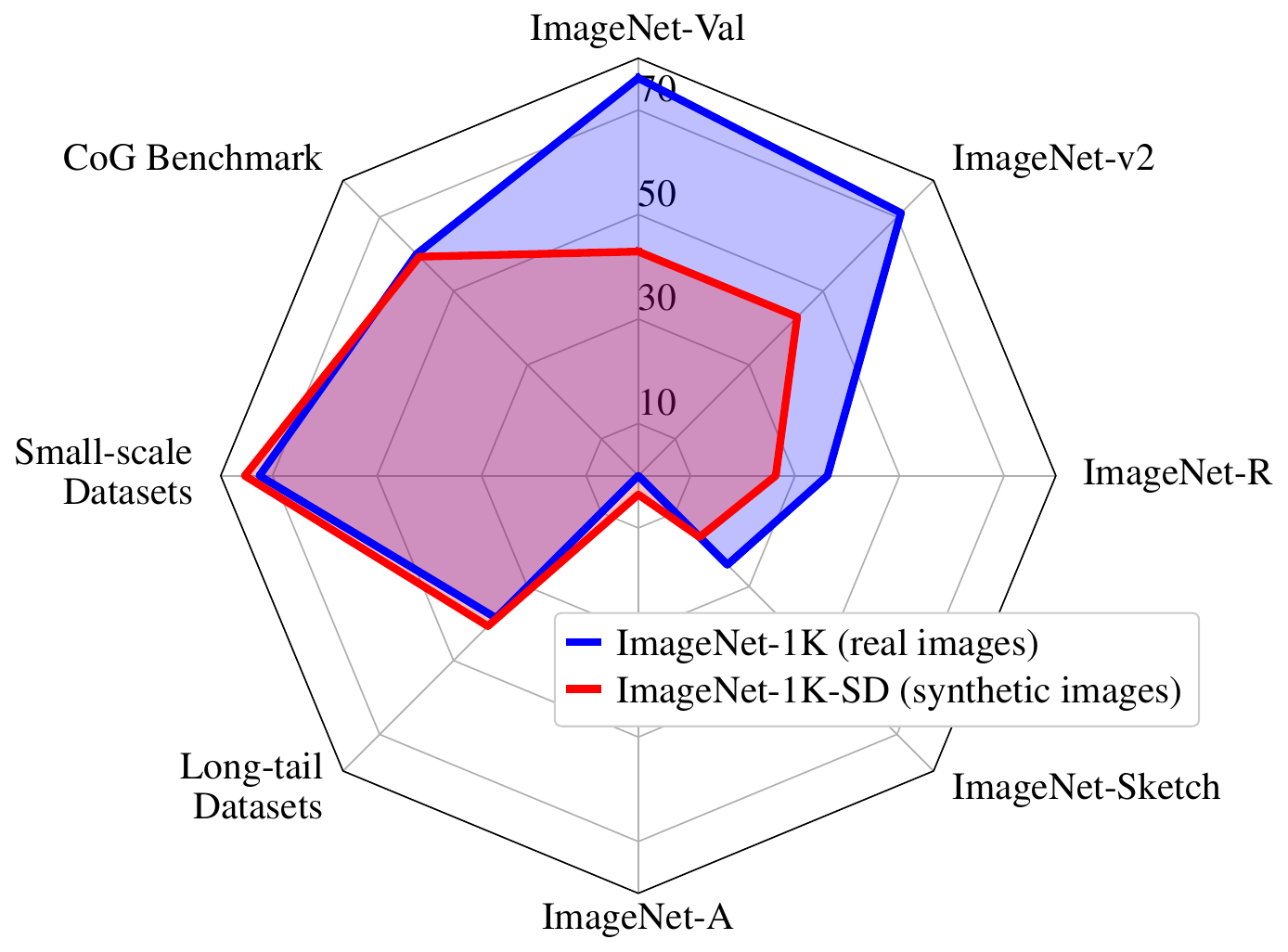}
        \caption{Top-1 accuracy, training on {\bf \imnetlong}.}
        \label{fig:spider_in1k_top1}
    \end{subfigure}
    ~
    \begin{subfigure}{0.46\linewidth}
        \centering
        \includegraphics[width=\linewidth]{res/spider/spider_in1k_top5.pdf}
        \caption{Top-5 accuracy, training on {\bf \imnetlong} (top-1 for transfer tasks).}
        \label{fig:spider_in1k_top5}
    \end{subfigure}
    \caption{
        {\bf Performance card of models} trained on either real or synthetic data for 100 classes of \imnethlong (\Cref{fig:spider_in100_top1,fig:spider_in100_top5}) and for all the 1000 classes of \imnetlong{} (\Cref{fig:spider_in1k_top1,fig:spider_in1k_top5}).
        In all figures, the {{\color{Blue}{blue polygon}}} shows the performance of a model trained on the real images from ImageNet, and the {{\color{Red}{red polygon}}} depicts the performance of a model trained \textit{only on synthetic data}, generated with Stable Diffusion~\cite{rombach2022high} using $p_c=$ ``$c$, $h_c$ inside $b$'' as the prompt.
        In~\Cref{fig:spider_in100_top1,fig:spider_in1k_top1} and in~\Cref{fig:spider_in100_top5,fig:spider_in1k_top5} we report top-1 and top-5 accuracy over the ImageNet datasets (\ie, ImageNet-Val/v2/R/A/Sketch), whereas, in all figures we report top-1 accuracy averaged over 8 transfer datasets.
        Note that~\Cref{fig:spider_in1k_top5} corresponds to Fig~1 of the main paper.
    }
    \label{fig:four_spiders}
\end{figure*}

%% file: fig/sd_params.tex
\begin{figure}[t!]
    \centering
    \begin{subfigure}[t]{\linewidth}
        \centering

        \includegraphics[width=\columnwidth]{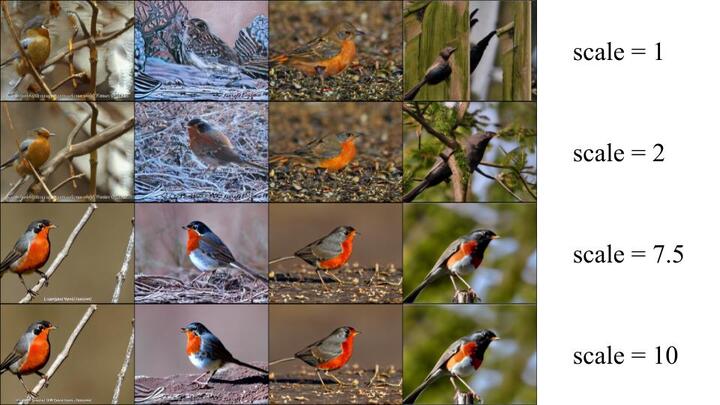} \vspace{-13.5pt}

        \includegraphics[width=\columnwidth]{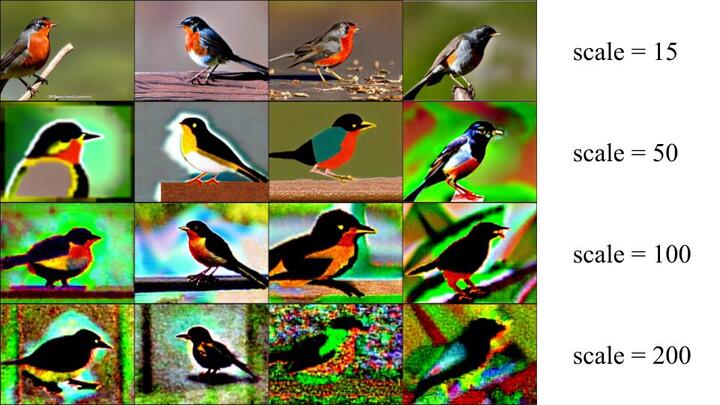}
        \caption{Varying the guidance scale parameter (steps = 50)}
        \label{fig:sd_scale}
        \end{subfigure}
        \vspace{20pt}

        \begin{subfigure}[t]{\linewidth}
        \centering
        \resizebox{\linewidth}{!}{
            \includegraphics[width=\columnwidth]{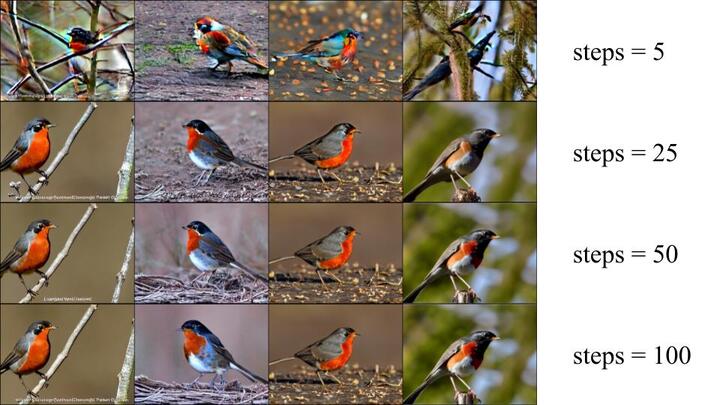}
        }
        \caption{Varying the number of diffusion steps (scale = 7.5)}
                \label{fig:sd_steps}
        \end{subfigure}
    \caption{\textbf{Qualitative results as we change the guidance scale parameter and the number of diffusion steps during Stable Diffusion generation.}
    The seed is fixed to \texttt{1947262} and the prompt is ``robin, American robin, Turdus migratorius''.
    Unless otherwise stated the scale (resp. steps) parameters are set to 7.5 (resp. 50).}
    \label{fig:sd_params}
\end{figure}

%% file: fig/shih-tzu.tex
\begin{figure*}[t]
    \centering
    \begin{subfigure}[t]{.80\linewidth}
        \centering
        \includegraphics[width=\linewidth]{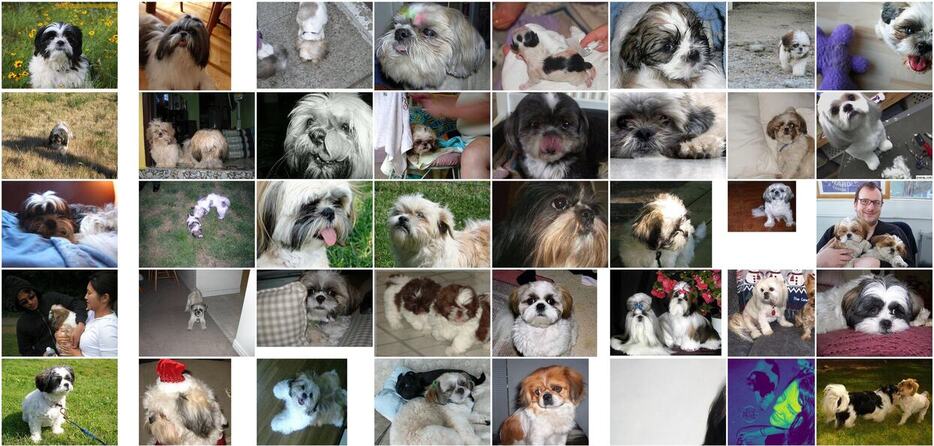}
        \caption{Real images from \imnetlong for class ``Shih-Tzu''}
        \label{fig:shih_tsu_real}
    \end{subfigure}
    \begin{subfigure}[t]{.80\linewidth}
        \centering
        \includegraphics[width=\linewidth]{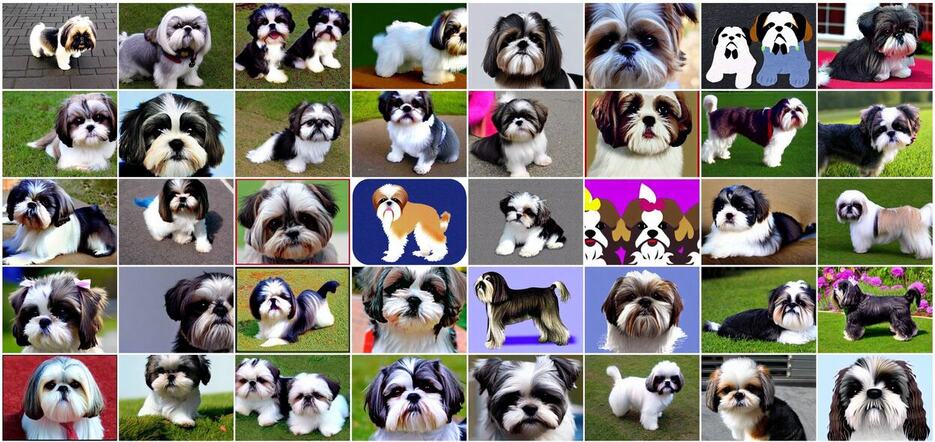}
        \caption{Synthetic images with prompt $p_c = $~``$c$'' for class ``Shih-Tzu''}
        \label{fig:shih_tsu_base}
    \end{subfigure}
    \begin{subfigure}[t]{.80\linewidth}
        \centering
        \includegraphics[width=\linewidth]{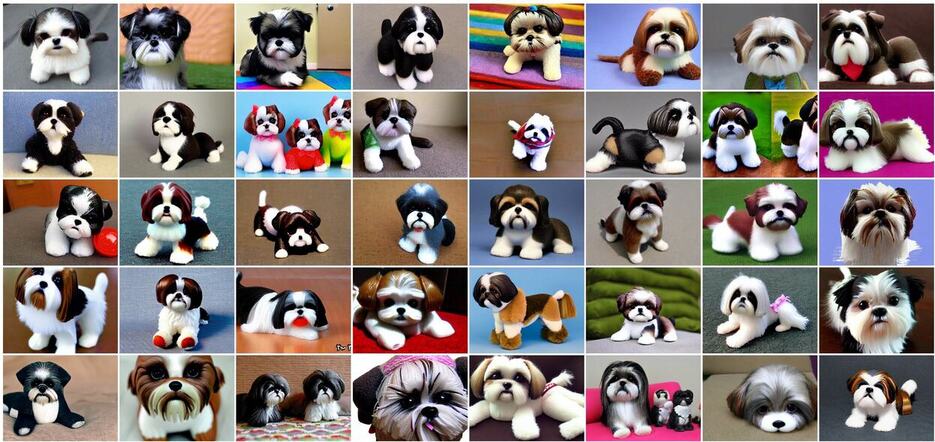}
        \caption{Synthetic images with prompt $p_c = $~``$c$, $h_c$'' for class ``Shih-Tzu''}
        \label{fig:shih_tsu_p}
    \end{subfigure}
    \caption{\textbf{Qualitative results for class ``Shih-Tzu''} to illustrate domain and diversity issues. Guidance scale is equal to 7.5.}
    \label{fig:shih_tsu}
\end{figure*}

\begin{figure*}[t]
    \centering
    \begin{subfigure}[t]{.80\linewidth}
        \centering
        \includegraphics[width=\columnwidth]{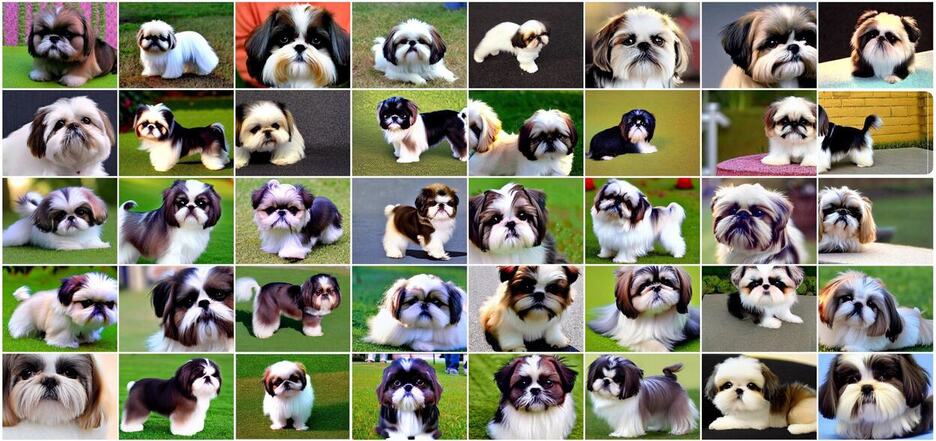}
        \caption{(cont.) Synthetic images with prompt $p_c = $~``$c$, $d_c$'' for class ``Shih-Tzu''}
        \label{fig:shih_tsu_d}
    \end{subfigure}
    \centering
    \centering
    \begin{subfigure}[t]{.80\linewidth}
        \centering
        \includegraphics[width=\columnwidth]{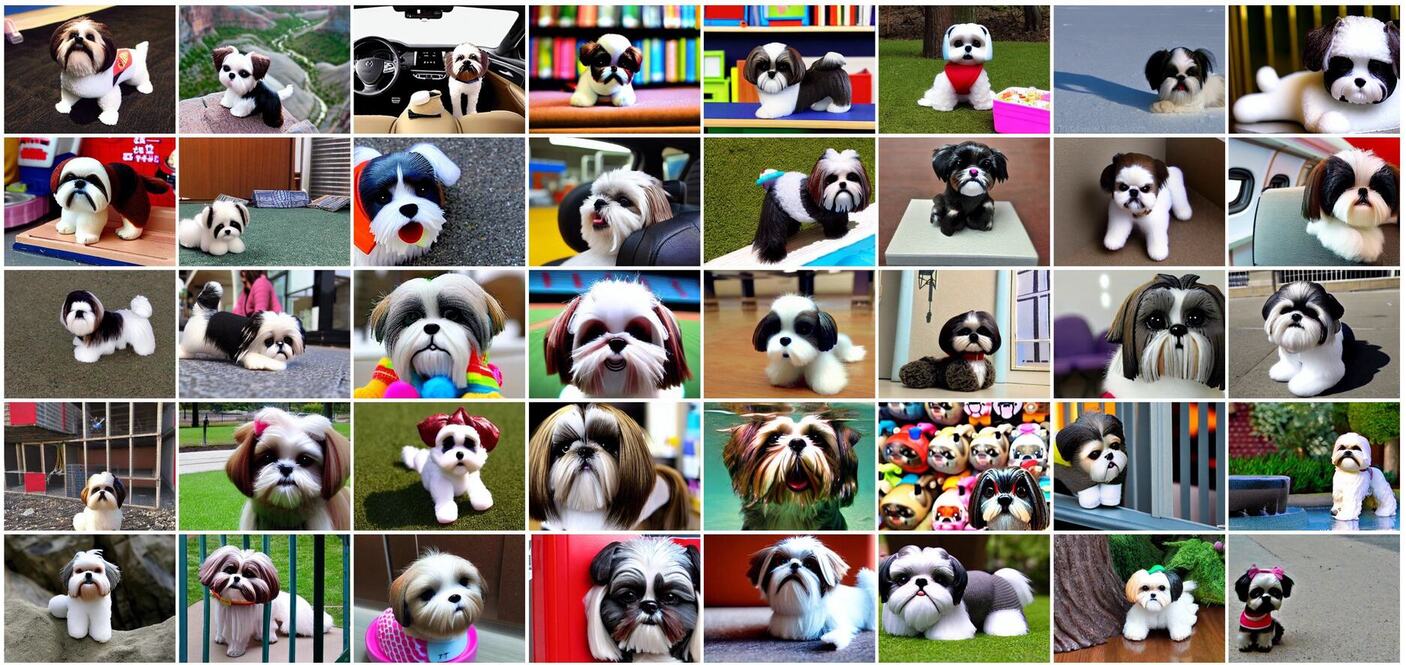}
        \caption{Synthetic images with prompt $p_c = $~``$c$, $h_c$ inside $b$''}
        \label{fig:shih_tsu_bg}
    \end{subfigure}
    \caption{\textbf{(cont.) Qualitative results for class ``Shih-Tzu'' to illustrate domain and diversity issues.}}
    \label{fig:shih_tsu}
\end{figure*}

%% file: fig/crabs.tex
\begin{figure*}[t]
    \centering
    \begin{subfigure}[t]{.96\linewidth}
        \centering
        \includegraphics[width=.46\linewidth]{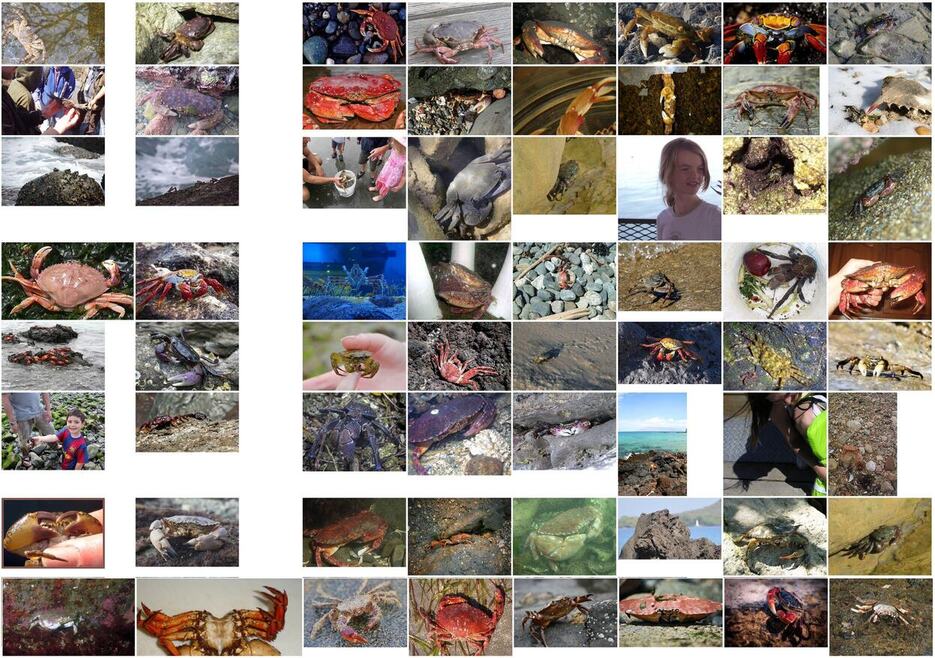} \hfill
        \includegraphics[width=.46\linewidth]{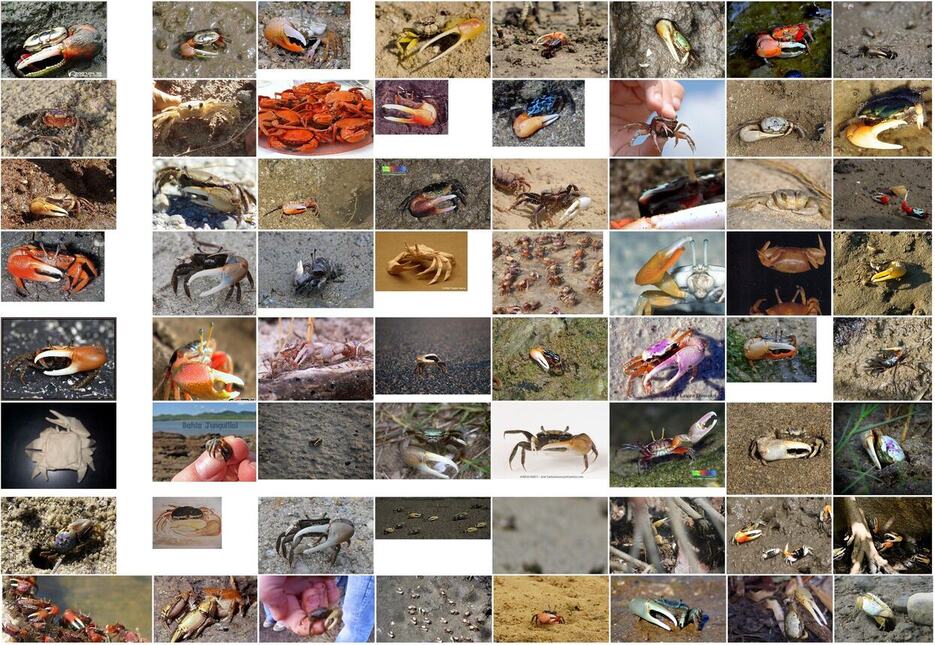}
        \caption{Real images from \imnetlong for classes ``Rock crab'' (left) and ``Fiddler crab'' (right)}
        \label{fig:crabs_real}
    \end{subfigure}
    \begin{subfigure}[t]{.96\linewidth}
        \centering
        \includegraphics[width=.46\linewidth]{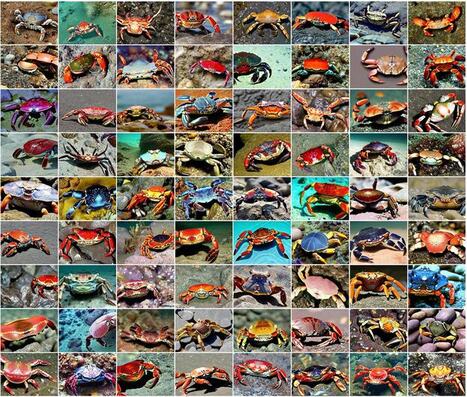} \hfill
        \includegraphics[width=.46\linewidth]{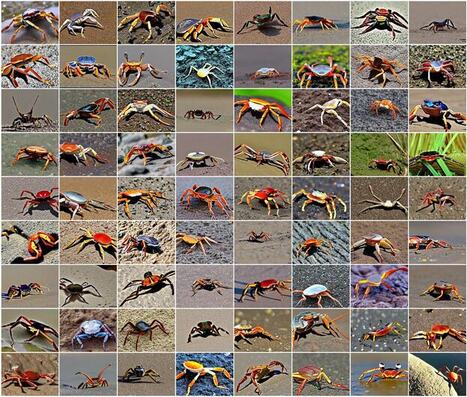}
        \caption{Synthetic images with prompt $p_c = $~``$c$'' for classes ``Rock crab'' (left) and ``Fiddler crab'' (right)}
        \label{fig:crabs_base}
    \end{subfigure}
    \begin{subfigure}[t]{.96\linewidth}
        \centering
        \includegraphics[width=.46\linewidth]{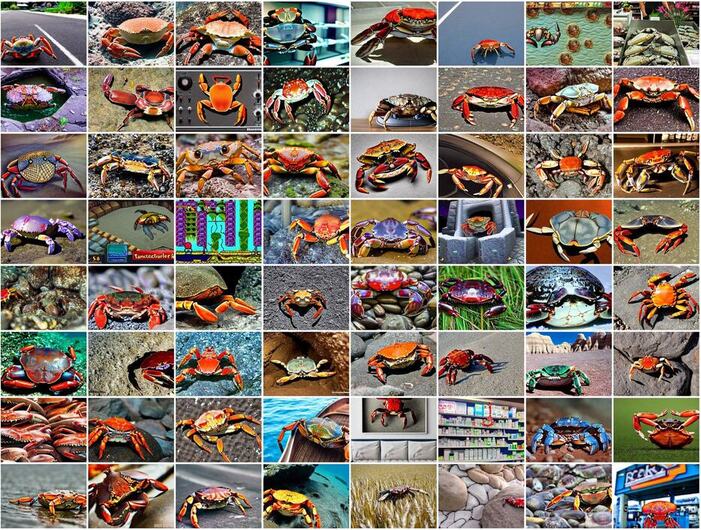} \hfill
        \includegraphics[width=.46\linewidth]{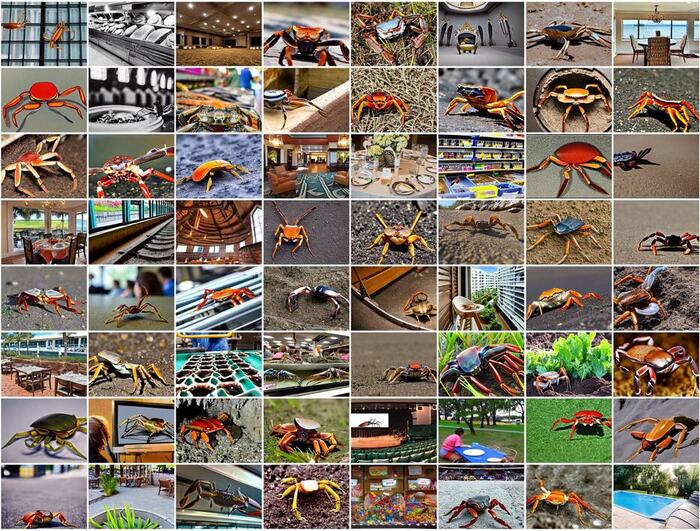}
        \caption{Synthetic images with prompt $p_c = $~``$c$, $h_c$ inside $b$'' for classes ``Rock crab'' (left) and ``Fiddler crab'' (right)}
        \label{fig:crabs_bg}
    \end{subfigure}
    \caption{\textbf{Qualitative results for classes ``Rock crab'' (left) and ``Fiddler crab'' (right)}, to illustrate issues around fine-grained and domain specific semantics.
    Guidance scale is equal to 7.5.}
    \label{fig:crabs}
\end{figure*}

%% file: fig/supp_qualitative_per_class.tex
{
\newcommand{\tabfigure}[2]{\raisebox{-.5\height}{\includegraphics[width=\linewidth]{#2}}}
\newcolumntype{L}{>{\centering\arraybackslash}m{0.3\linewidth}}

\begin{figure*}[t]
    \centering
    \adjustbox{max width=\linewidth, totalheight=0.97\textheight}{
    \begin{tabular}{p{0.1\linewidth}  L L L}
        \toprule
        Synset & {\centering real images} & $p_c=$ ``$c$'' & $p_c=$ ``$c$, $h_c$ inside $b$'' \\
               &                          & guidance scale 7.5 & guidance scale 2 \\
        \toprule
robin & \tabfigure{width=0.3\linewidth}{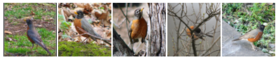} & \tabfigure{width=0.3\linewidth}{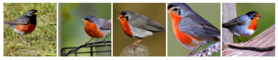} & \tabfigure{width=0.3\linewidth}{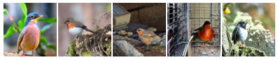} \\
Gila monster & \tabfigure{width=0.3\linewidth}{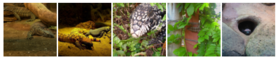} & \tabfigure{width=0.3\linewidth}{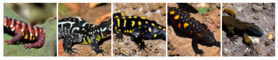} & \tabfigure{width=0.3\linewidth}{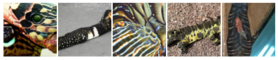} \\
hognose snake & \tabfigure{width=0.3\linewidth}{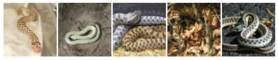} & \tabfigure{width=0.3\linewidth}{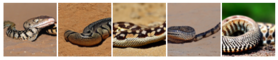} & \tabfigure{width=0.3\linewidth}{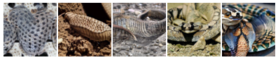} \\
garter snake & \tabfigure{width=0.3\linewidth}{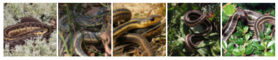} & \tabfigure{width=0.3\linewidth}{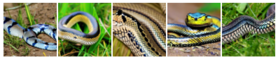} & \tabfigure{width=0.3\linewidth}{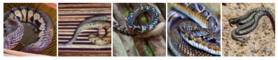} \\
green mamba & \tabfigure{width=0.3\linewidth}{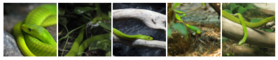} & \tabfigure{width=0.3\linewidth}{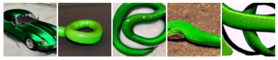} & \tabfigure{width=0.3\linewidth}{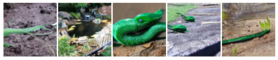} \\
garden spider & \tabfigure{width=0.3\linewidth}{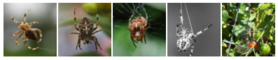} & \tabfigure{width=0.3\linewidth}{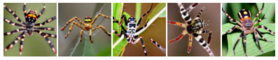} & \tabfigure{width=0.3\linewidth}{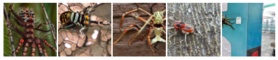} \\
lorikeet & \tabfigure{width=0.3\linewidth}{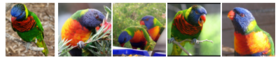} & \tabfigure{width=0.3\linewidth}{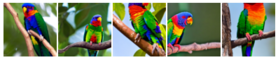} & \tabfigure{width=0.3\linewidth}{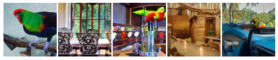} \\
goose & \tabfigure{width=0.3\linewidth}{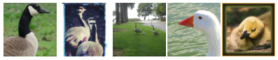} & \tabfigure{width=0.3\linewidth}{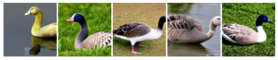} & \tabfigure{width=0.3\linewidth}{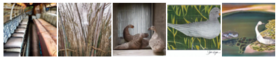} \\
rock crab & \tabfigure{width=0.3\linewidth}{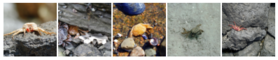} & \tabfigure{width=0.3\linewidth}{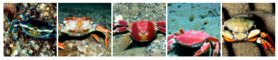} & \tabfigure{width=0.3\linewidth}{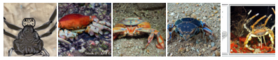} \\
fiddler crab & \tabfigure{width=0.3\linewidth}{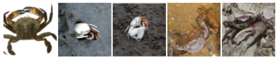} & \tabfigure{width=0.3\linewidth}{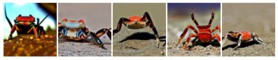} & \tabfigure{width=0.3\linewidth}{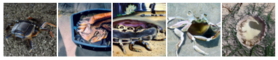} \\
American lobster & \tabfigure{width=0.3\linewidth}{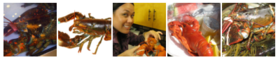} & \tabfigure{width=0.3\linewidth}{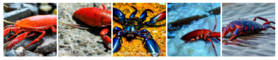} & \tabfigure{width=0.3\linewidth}{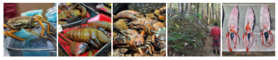} \\
little blue heron & \tabfigure{width=0.3\linewidth}{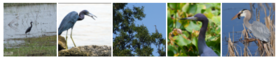} & \tabfigure{width=0.3\linewidth}{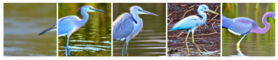} & \tabfigure{width=0.3\linewidth}{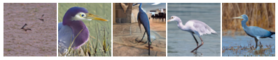} \\
American coot & \tabfigure{width=0.3\linewidth}{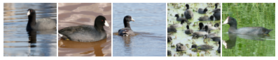} & \tabfigure{width=0.3\linewidth}{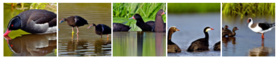} & \tabfigure{width=0.3\linewidth}{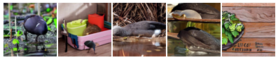} \\
Chihuahua & \tabfigure{width=0.3\linewidth}{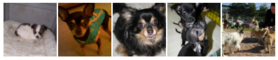} & \tabfigure{width=0.3\linewidth}{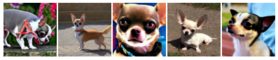} & \tabfigure{width=0.3\linewidth}{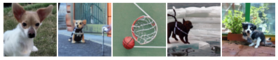} \\
Shih-Tzu & \tabfigure{width=0.3\linewidth}{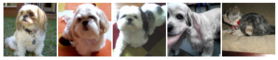} & \tabfigure{width=0.3\linewidth}{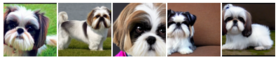} & \tabfigure{width=0.3\linewidth}{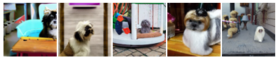} \\
papillon & \tabfigure{width=0.3\linewidth}{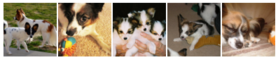} & \tabfigure{width=0.3\linewidth}{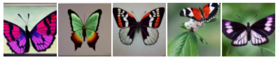} & \tabfigure{width=0.3\linewidth}{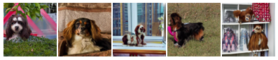} \\
toy terrier & \tabfigure{width=0.3\linewidth}{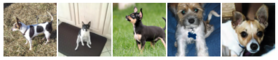} & \tabfigure{width=0.3\linewidth}{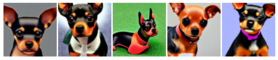} & \tabfigure{width=0.3\linewidth}{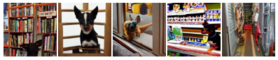} \\
Walker hound & \tabfigure{width=0.3\linewidth}{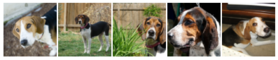} & \tabfigure{width=0.3\linewidth}{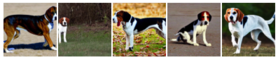} & \tabfigure{width=0.3\linewidth}{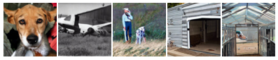} \\
English foxhound & \tabfigure{width=0.3\linewidth}{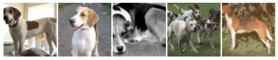} & \tabfigure{width=0.3\linewidth}{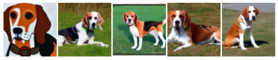} & \tabfigure{width=0.3\linewidth}{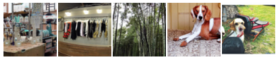} \\
borzoi & \tabfigure{width=0.3\linewidth}{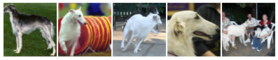} & \tabfigure{width=0.3\linewidth}{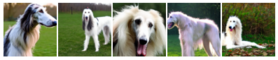} & \tabfigure{width=0.3\linewidth}{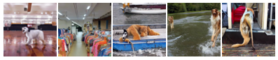} \\
\bottomrule
    \end{tabular}
    }
    \caption{
        {\bf Visualization of the 100 \imnethlong classes} for the three different datasets:
        \imnethlong-Val (real) and two \imnethsdlong datasets created with prompts $p_c=$ ``$c$'' and $p_c=$ ``$c$, $h_c$ inside $b$''.
    }
    \label{fig:images_for_im100_classes}
\end{figure*}

\begin{figure*}
    \centering
    \adjustbox{max width=\linewidth, totalheight=0.97\textheight}{
    \begin{tabular}{p{0.1\linewidth}  L L L}
        \toprule
        Synset & real images & $p_c=$ ``$c$'' & $p_c=$ ``$c$, $h_c$ inside $b$'' \\
               &                          & guidance scale 7.5 & guidance scale 2 \\
        \toprule
Saluki & \tabfigure{width=0.3\linewidth}{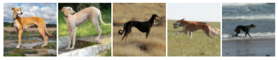} & \tabfigure{width=0.3\linewidth}{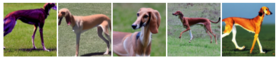} & \tabfigure{width=0.3\linewidth}{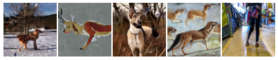} \\
American Staffordshire terrier & \tabfigure{width=0.3\linewidth}{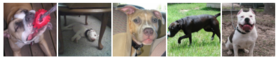} & \tabfigure{width=0.3\linewidth}{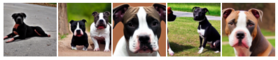} & \tabfigure{width=0.3\linewidth}{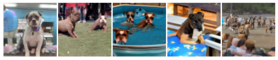} \\
Chesapeake Bay retriever & \tabfigure{width=0.3\linewidth}{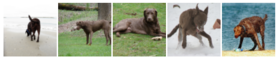} & \tabfigure{width=0.3\linewidth}{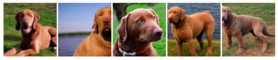} & \tabfigure{width=0.3\linewidth}{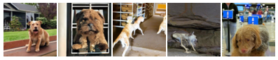} \\
vizsla & \tabfigure{width=0.3\linewidth}{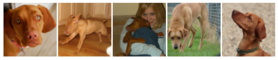} & \tabfigure{width=0.3\linewidth}{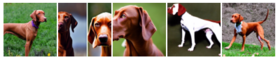} & \tabfigure{width=0.3\linewidth}{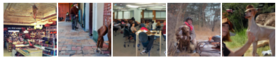} \\
kuvasz & \tabfigure{width=0.3\linewidth}{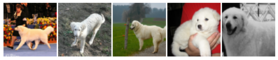} & \tabfigure{width=0.3\linewidth}{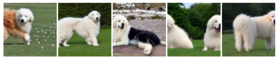} & \tabfigure{width=0.3\linewidth}{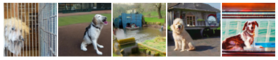} \\
komondor & \tabfigure{width=0.3\linewidth}{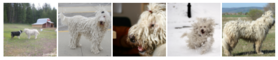} & \tabfigure{width=0.3\linewidth}{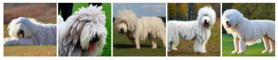} & \tabfigure{width=0.3\linewidth}{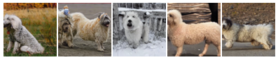} \\
Rottweiler & \tabfigure{width=0.3\linewidth}{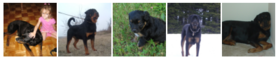} & \tabfigure{width=0.3\linewidth}{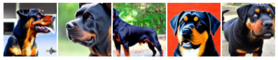} & \tabfigure{width=0.3\linewidth}{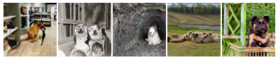} \\
Doberman & \tabfigure{width=0.3\linewidth}{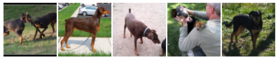} & \tabfigure{width=0.3\linewidth}{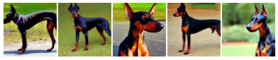} & \tabfigure{width=0.3\linewidth}{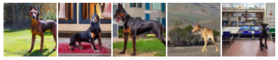} \\
boxer & \tabfigure{width=0.3\linewidth}{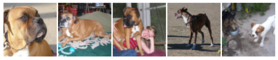} & \tabfigure{width=0.3\linewidth}{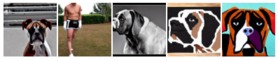} & \tabfigure{width=0.3\linewidth}{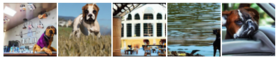} \\
Great Dane & \tabfigure{width=0.3\linewidth}{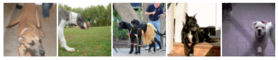} & \tabfigure{width=0.3\linewidth}{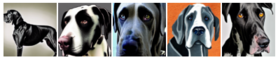} & \tabfigure{width=0.3\linewidth}{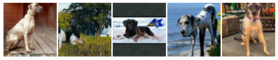} \\
standard poodle & \tabfigure{width=0.3\linewidth}{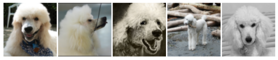} & \tabfigure{width=0.3\linewidth}{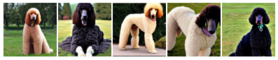} & \tabfigure{width=0.3\linewidth}{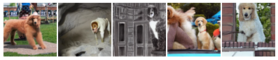} \\
Mexican hairless & \tabfigure{width=0.3\linewidth}{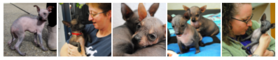} & \tabfigure{width=0.3\linewidth}{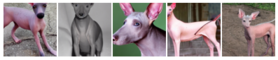} & \tabfigure{width=0.3\linewidth}{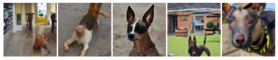} \\
coyote & \tabfigure{width=0.3\linewidth}{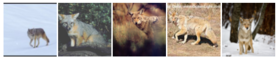} & \tabfigure{width=0.3\linewidth}{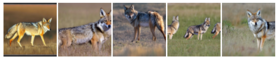} & \tabfigure{width=0.3\linewidth}{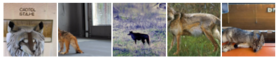} \\
African hunting dog & \tabfigure{width=0.3\linewidth}{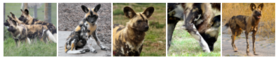} & \tabfigure{width=0.3\linewidth}{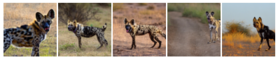} & \tabfigure{width=0.3\linewidth}{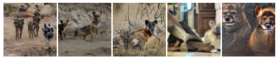} \\
red fox & \tabfigure{width=0.3\linewidth}{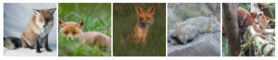} & \tabfigure{width=0.3\linewidth}{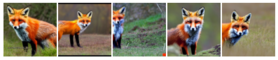} & \tabfigure{width=0.3\linewidth}{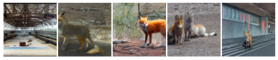} \\
tabby & \tabfigure{width=0.3\linewidth}{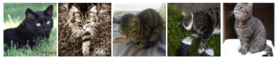} & \tabfigure{width=0.3\linewidth}{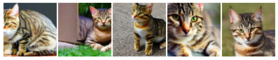} & \tabfigure{width=0.3\linewidth}{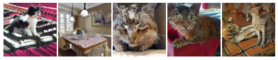} \\
meerkat & \tabfigure{width=0.3\linewidth}{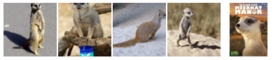} & \tabfigure{width=0.3\linewidth}{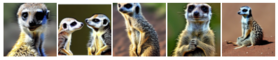} & \tabfigure{width=0.3\linewidth}{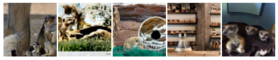} \\
dung beetle & \tabfigure{width=0.3\linewidth}{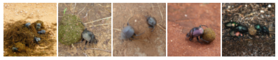} & \tabfigure{width=0.3\linewidth}{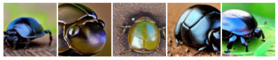} & \tabfigure{width=0.3\linewidth}{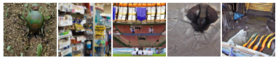} \\
walking stick & \tabfigure{width=0.3\linewidth}{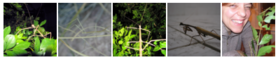} & \tabfigure{width=0.3\linewidth}{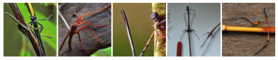} & \tabfigure{width=0.3\linewidth}{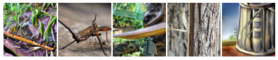} \\
leafhopper & \tabfigure{width=0.3\linewidth}{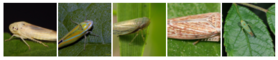} & \tabfigure{width=0.3\linewidth}{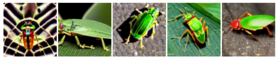} & \tabfigure{width=0.3\linewidth}{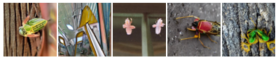} \\
\bottomrule
    \end{tabular}
    }
    \caption{
        {\bf (cont.) Visualization of the 100 \imnethlong classes} for the three different datasets:
        \imnethlong-Val (real) and two \imnethsdlong datasets created with prompts $p_c=$ ``$c$'' and $p_c=$ ``$c$, $h_c$ inside $b$''.
    }
\end{figure*}

\begin{figure*}
    \centering
    \adjustbox{max width=\linewidth, totalheight=0.97\textheight}{
    \begin{tabular}{p{0.1\linewidth}  L L L}
        \toprule
        Synset & real images & $p_c=$ ``$c$'' & $p_c=$ ``$c$, $h_c$ inside $b$'' \\
               &                          & guidance scale 7.5 & guidance scale 2 \\
        \toprule
hare & \tabfigure{width=0.3\linewidth}{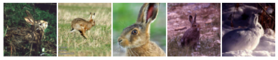} & \tabfigure{width=0.3\linewidth}{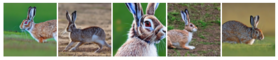} & \tabfigure{width=0.3\linewidth}{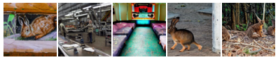} \\
wild boar & \tabfigure{width=0.3\linewidth}{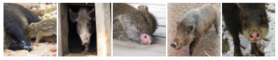} & \tabfigure{width=0.3\linewidth}{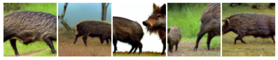} & \tabfigure{width=0.3\linewidth}{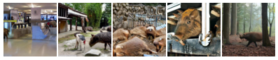} \\
gibbon & \tabfigure{width=0.3\linewidth}{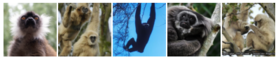} & \tabfigure{width=0.3\linewidth}{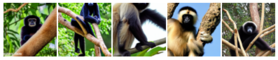} & \tabfigure{width=0.3\linewidth}{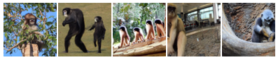} \\
langur & \tabfigure{width=0.3\linewidth}{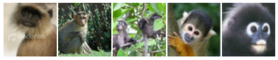} & \tabfigure{width=0.3\linewidth}{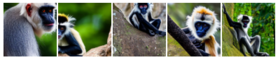} & \tabfigure{width=0.3\linewidth}{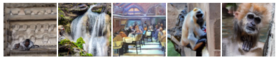} \\
ambulance & \tabfigure{width=0.3\linewidth}{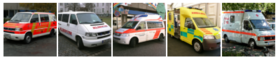} & \tabfigure{width=0.3\linewidth}{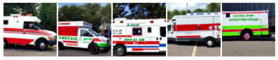} & \tabfigure{width=0.3\linewidth}{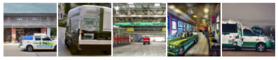} \\
bannister & \tabfigure{width=0.3\linewidth}{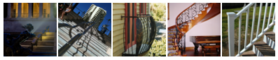} & \tabfigure{width=0.3\linewidth}{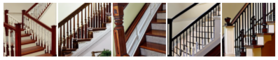} & \tabfigure{width=0.3\linewidth}{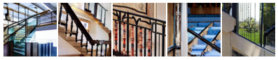} \\
bassinet & \tabfigure{width=0.3\linewidth}{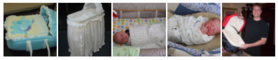} & \tabfigure{width=0.3\linewidth}{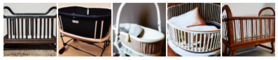} & \tabfigure{width=0.3\linewidth}{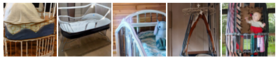} \\
boathouse & \tabfigure{width=0.3\linewidth}{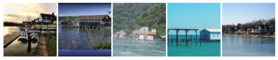} & \tabfigure{width=0.3\linewidth}{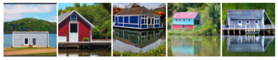} & \tabfigure{width=0.3\linewidth}{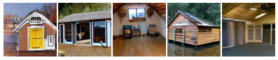} \\
bonnet & \tabfigure{width=0.3\linewidth}{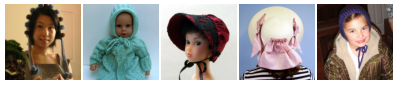} & \tabfigure{width=0.3\linewidth}{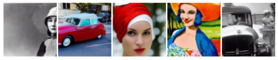} & \tabfigure{width=0.3\linewidth}{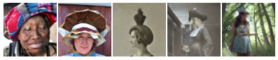} \\
bottlecap & \tabfigure{width=0.3\linewidth}{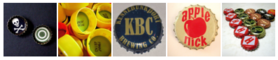} & \tabfigure{width=0.3\linewidth}{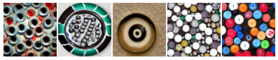} & \tabfigure{width=0.3\linewidth}{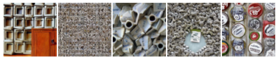} \\
car wheel & \tabfigure{width=0.3\linewidth}{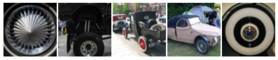} & \tabfigure{width=0.3\linewidth}{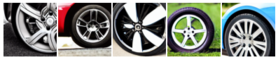} & \tabfigure{width=0.3\linewidth}{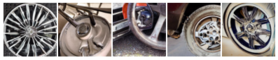} \\
chime & \tabfigure{width=0.3\linewidth}{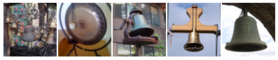} & \tabfigure{width=0.3\linewidth}{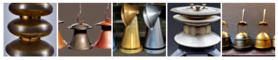} & \tabfigure{width=0.3\linewidth}{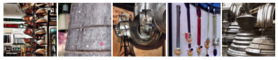} \\
cinema & \tabfigure{width=0.3\linewidth}{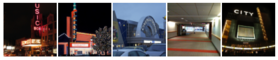} & \tabfigure{width=0.3\linewidth}{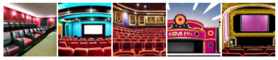} & \tabfigure{width=0.3\linewidth}{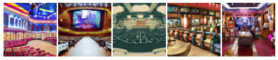} \\
cocktail shaker & \tabfigure{width=0.3\linewidth}{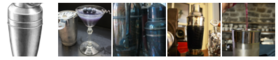} & \tabfigure{width=0.3\linewidth}{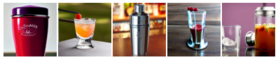} & \tabfigure{width=0.3\linewidth}{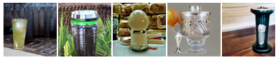} \\
computer keyboard & \tabfigure{width=0.3\linewidth}{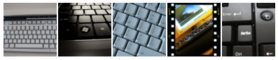} & \tabfigure{width=0.3\linewidth}{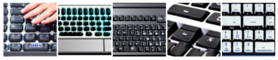} & \tabfigure{width=0.3\linewidth}{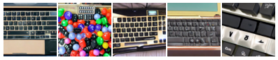} \\
Dutch oven & \tabfigure{width=0.3\linewidth}{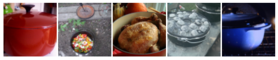} & \tabfigure{width=0.3\linewidth}{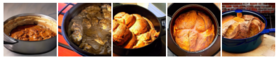} & \tabfigure{width=0.3\linewidth}{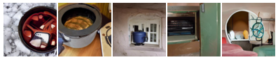} \\
football helmet & \tabfigure{width=0.3\linewidth}{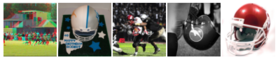} & \tabfigure{width=0.3\linewidth}{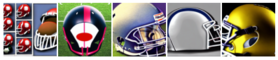} & \tabfigure{width=0.3\linewidth}{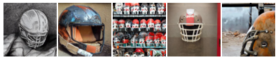} \\
gasmask & \tabfigure{width=0.3\linewidth}{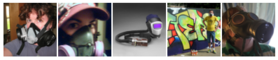} & \tabfigure{width=0.3\linewidth}{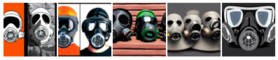} & \tabfigure{width=0.3\linewidth}{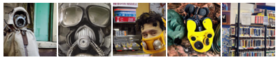} \\
hard disc & \tabfigure{width=0.3\linewidth}{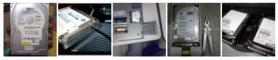} & \tabfigure{width=0.3\linewidth}{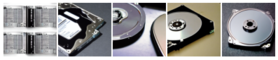} & \tabfigure{width=0.3\linewidth}{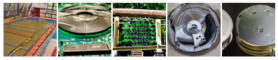} \\
harmonica & \tabfigure{width=0.3\linewidth}{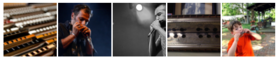} & \tabfigure{width=0.3\linewidth}{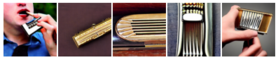} & \tabfigure{width=0.3\linewidth}{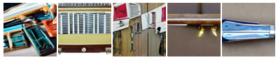} \\
\bottomrule
    \end{tabular}
    }
    \caption{
        {\bf (cont.) Visualization of the images for the 100 \imnethlong classes} in the three different datasets:
        \imnethlong-Val (real) and two \imnethsdlong datasets created with prompts $p_c=$ ``$c$'' and $p_c=$ ``$c$, $h_c$ inside $b$''.
    }
\end{figure*}

\begin{figure*}
    \centering
    \adjustbox{max width=\linewidth, totalheight=0.97\textheight}{
    \begin{tabular}{p{0.1\linewidth}  L L L}
        \toprule
        Synset & real images & $p_c=$ ``$c$'' & $p_c=$ ``$c$, $h_c$ inside $b$'' \\
               &                          & guidance scale 7.5 & guidance scale 2 \\
        \toprule
honeycomb & \tabfigure{width=0.3\linewidth}{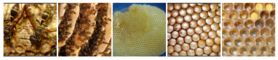} & \tabfigure{width=0.3\linewidth}{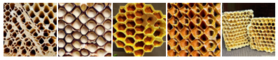} & \tabfigure{width=0.3\linewidth}{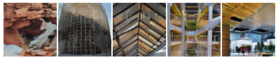} \\
iron & \tabfigure{width=0.3\linewidth}{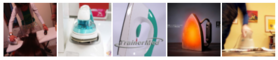} & \tabfigure{width=0.3\linewidth}{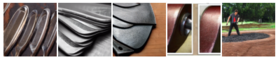} & \tabfigure{width=0.3\linewidth}{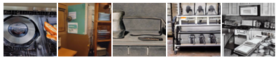} \\
jean & \tabfigure{width=0.3\linewidth}{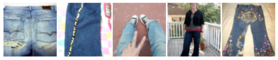} & \tabfigure{width=0.3\linewidth}{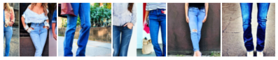} & \tabfigure{width=0.3\linewidth}{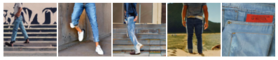} \\
lampshade & \tabfigure{width=0.3\linewidth}{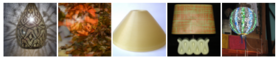} & \tabfigure{width=0.3\linewidth}{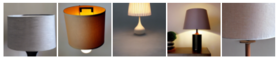} & \tabfigure{width=0.3\linewidth}{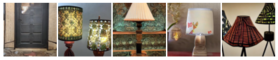} \\
laptop & \tabfigure{width=0.3\linewidth}{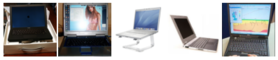} & \tabfigure{width=0.3\linewidth}{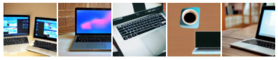} & \tabfigure{width=0.3\linewidth}{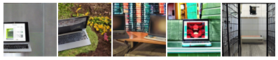} \\
milk can & \tabfigure{width=0.3\linewidth}{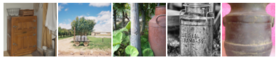} & \tabfigure{width=0.3\linewidth}{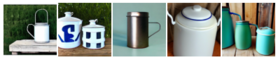} & \tabfigure{width=0.3\linewidth}{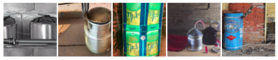} \\
mixing bowl & \tabfigure{width=0.3\linewidth}{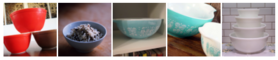} & \tabfigure{width=0.3\linewidth}{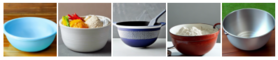} & \tabfigure{width=0.3\linewidth}{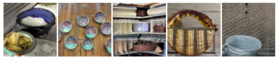} \\
modem & \tabfigure{width=0.3\linewidth}{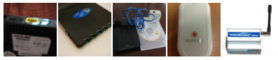} & \tabfigure{width=0.3\linewidth}{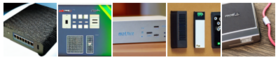} & \tabfigure{width=0.3\linewidth}{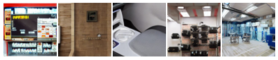} \\
moped & \tabfigure{width=0.3\linewidth}{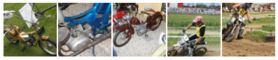} & \tabfigure{width=0.3\linewidth}{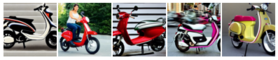} & \tabfigure{width=0.3\linewidth}{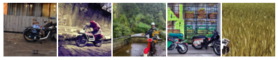} \\
mortarboard & \tabfigure{width=0.3\linewidth}{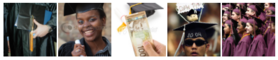} & \tabfigure{width=0.3\linewidth}{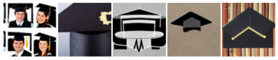} & \tabfigure{width=0.3\linewidth}{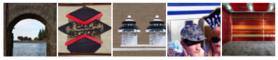} \\
mousetrap & \tabfigure{width=0.3\linewidth}{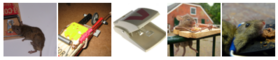} & \tabfigure{width=0.3\linewidth}{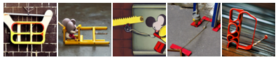} & \tabfigure{width=0.3\linewidth}{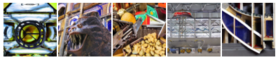} \\
obelisk & \tabfigure{width=0.3\linewidth}{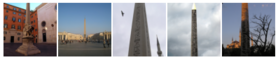} & \tabfigure{width=0.3\linewidth}{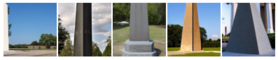} & \tabfigure{width=0.3\linewidth}{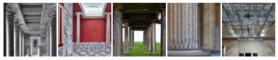} \\
park bench & \tabfigure{width=0.3\linewidth}{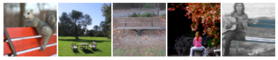} & \tabfigure{width=0.3\linewidth}{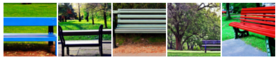} & \tabfigure{width=0.3\linewidth}{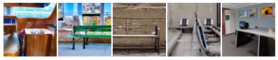} \\
pedestal & \tabfigure{width=0.3\linewidth}{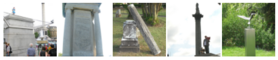} & \tabfigure{width=0.3\linewidth}{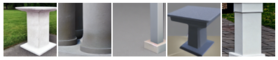} & \tabfigure{width=0.3\linewidth}{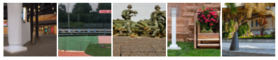} \\
pickup & \tabfigure{width=0.3\linewidth}{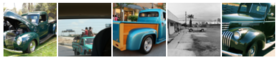} & \tabfigure{width=0.3\linewidth}{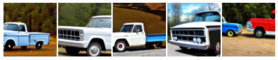} & \tabfigure{width=0.3\linewidth}{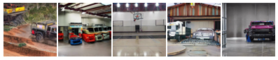} \\
pirate & \tabfigure{width=0.3\linewidth}{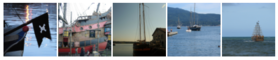} & \tabfigure{width=0.3\linewidth}{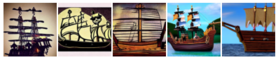} & \tabfigure{width=0.3\linewidth}{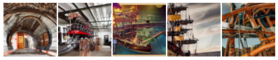} \\
purse & \tabfigure{width=0.3\linewidth}{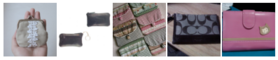} & \tabfigure{width=0.3\linewidth}{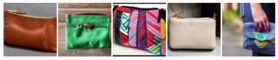} & \tabfigure{width=0.3\linewidth}{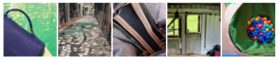} \\
reel & \tabfigure{width=0.3\linewidth}{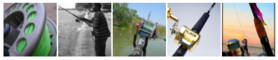} & \tabfigure{width=0.3\linewidth}{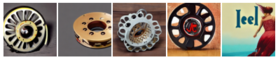} & \tabfigure{width=0.3\linewidth}{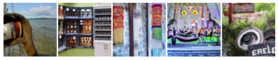} \\
rocking chair & \tabfigure{width=0.3\linewidth}{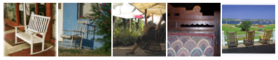} & \tabfigure{width=0.3\linewidth}{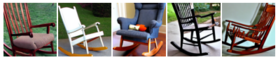} & \tabfigure{width=0.3\linewidth}{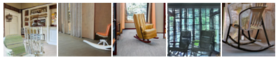} \\
rotisserie & \tabfigure{width=0.3\linewidth}{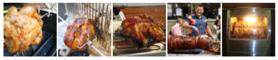} & \tabfigure{width=0.3\linewidth}{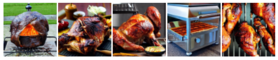} & \tabfigure{width=0.3\linewidth}{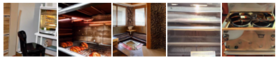} \\
\bottomrule
    \end{tabular}
    }
    \caption{
        {\bf (cont.) Visualization of the 100 \imnethlong classes} for the three different datasets:
        \imnethlong-Val (real) and two \imnethsdlong datasets created with prompts $p_c=$ ``$c$'' and $p_c=$ ``$c$, $h_c$ inside $b$''.
    }
\end{figure*}

\begin{figure*}
    \centering
    \adjustbox{max width=\linewidth, totalheight=0.97\textheight}{
    \begin{tabular}{p{0.1\linewidth}  L L L}
        \toprule
        Synset & real images & $p_c=$ ``$c$'' & $p_c=$ ``$c$, $h_c$ inside $b$'' \\
               &                          & guidance scale 7.5 & guidance scale 2 \\
        \toprule
safety pin & \tabfigure{width=0.3\linewidth}{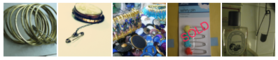} & \tabfigure{width=0.3\linewidth}{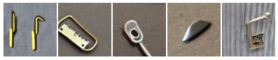} & \tabfigure{width=0.3\linewidth}{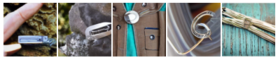} \\
sarong & \tabfigure{width=0.3\linewidth}{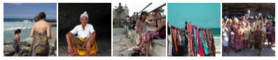} & \tabfigure{width=0.3\linewidth}{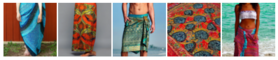} & \tabfigure{width=0.3\linewidth}{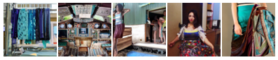} \\
ski mask & \tabfigure{width=0.3\linewidth}{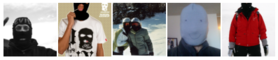} & \tabfigure{width=0.3\linewidth}{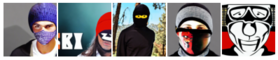} & \tabfigure{width=0.3\linewidth}{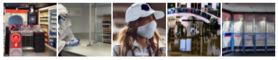} \\
slide rule & \tabfigure{width=0.3\linewidth}{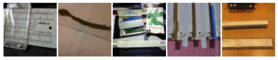} & \tabfigure{width=0.3\linewidth}{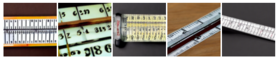} & \tabfigure{width=0.3\linewidth}{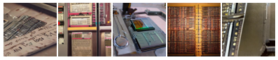} \\
stretcher & \tabfigure{width=0.3\linewidth}{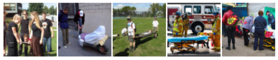} & \tabfigure{width=0.3\linewidth}{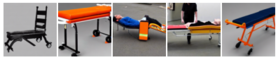} & \tabfigure{width=0.3\linewidth}{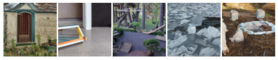} \\
theater curtain & \tabfigure{width=0.3\linewidth}{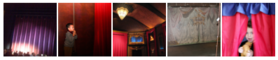} & \tabfigure{width=0.3\linewidth}{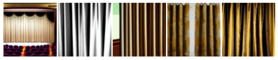} & \tabfigure{width=0.3\linewidth}{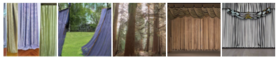} \\
throne & \tabfigure{width=0.3\linewidth}{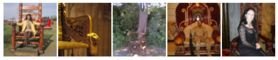} & \tabfigure{width=0.3\linewidth}{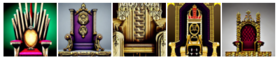} & \tabfigure{width=0.3\linewidth}{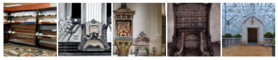} \\
tile roof & \tabfigure{width=0.3\linewidth}{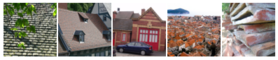} & \tabfigure{width=0.3\linewidth}{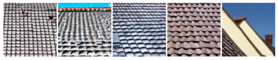} & \tabfigure{width=0.3\linewidth}{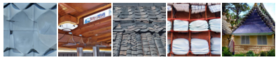} \\
tripod & \tabfigure{width=0.3\linewidth}{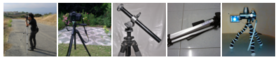} & \tabfigure{width=0.3\linewidth}{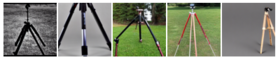} & \tabfigure{width=0.3\linewidth}{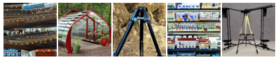} \\
tub & \tabfigure{width=0.3\linewidth}{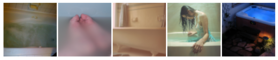} & \tabfigure{width=0.3\linewidth}{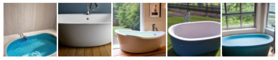} & \tabfigure{width=0.3\linewidth}{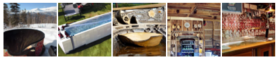} \\
vacuum & \tabfigure{width=0.3\linewidth}{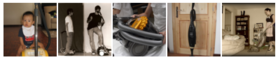} & \tabfigure{width=0.3\linewidth}{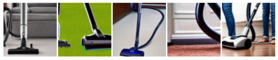} & \tabfigure{width=0.3\linewidth}{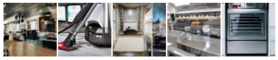} \\
window screen & \tabfigure{width=0.3\linewidth}{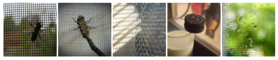} & \tabfigure{width=0.3\linewidth}{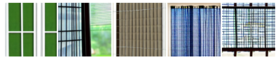} & \tabfigure{width=0.3\linewidth}{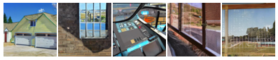} \\
wing & \tabfigure{width=0.3\linewidth}{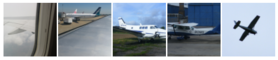} & \tabfigure{width=0.3\linewidth}{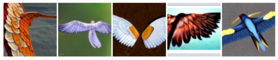} & \tabfigure{width=0.3\linewidth}{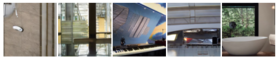} \\
head cabbage & \tabfigure{width=0.3\linewidth}{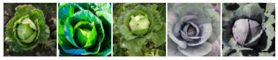} & \tabfigure{width=0.3\linewidth}{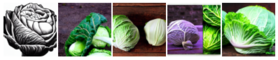} & \tabfigure{width=0.3\linewidth}{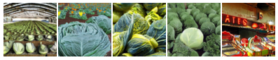} \\
cauliflower & \tabfigure{width=0.3\linewidth}{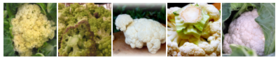} & \tabfigure{width=0.3\linewidth}{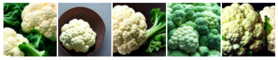} & \tabfigure{width=0.3\linewidth}{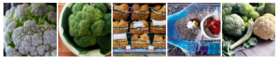} \\
pineapple & \tabfigure{width=0.3\linewidth}{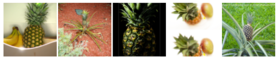} & \tabfigure{width=0.3\linewidth}{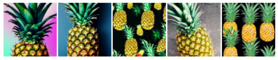} & \tabfigure{width=0.3\linewidth}{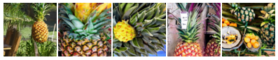} \\
carbonara & \tabfigure{width=0.3\linewidth}{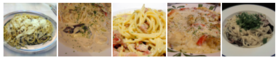} & \tabfigure{width=0.3\linewidth}{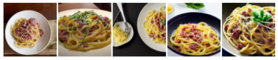} & \tabfigure{width=0.3\linewidth}{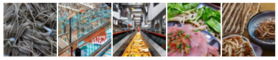} \\
chocolate sauce & \tabfigure{width=0.3\linewidth}{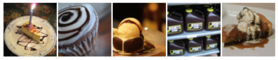} & \tabfigure{width=0.3\linewidth}{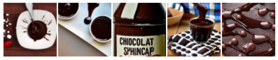} & \tabfigure{width=0.3\linewidth}{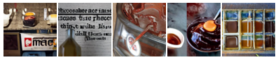} \\
gyromitra & \tabfigure{width=0.3\linewidth}{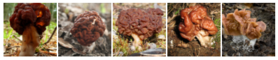} & \tabfigure{width=0.3\linewidth}{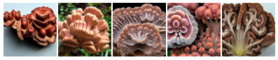} & \tabfigure{width=0.3\linewidth}{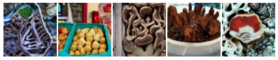} \\
stinkhorn & \tabfigure{width=0.3\linewidth}{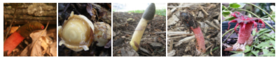} & \tabfigure{width=0.3\linewidth}{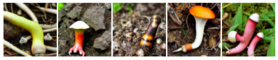} & \tabfigure{width=0.3\linewidth}{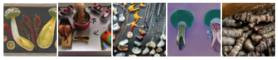} \\
\bottomrule
    \end{tabular}
    }
    \caption{
        {\bf (cont.) Visualization of the 100 \imnethlong classes} for the three different datasets:
        \imnethlong-Val (real) and two \imnethsdlong datasets created with prompts $p_c=$ ``$c$'' and $p_c=$ ``$c$, $h_c$ inside $b$''.
    }
\end{figure*}
}